\documentclass{article}

\usepackage{arxiv}

\usepackage[utf8]{inputenc} 
\usepackage[T1]{fontenc}    
\usepackage{hyperref}       
\usepackage{url}            
\usepackage{booktabs}       
\usepackage{amsfonts}       
\usepackage{nicefrac}       
\usepackage{microtype}      
\usepackage{lipsum}
\usepackage{graphicx}
\usepackage{amsmath}
\usepackage{float}
\usepackage{booktabs}
\usepackage{siunitx}
\usepackage{mathtools}
\usepackage[style=apa]{biblatex}
\usepackage{makecell}
\usepackage{algorithm}
\usepackage{algorithmic}
\usepackage{subcaption}
\usepackage{todonotes}

\graphicspath{ {./images/} }

\newif\ifanonymous
\anonymousfalse

\title{Thousand-Brains Systems: Sensorimotor Intelligence for Rapid, Robust Learning and Inference}

\ifanonymous
  \author{Anonymous Authors \\
  Paper under double-blind review}
\else
    \author{
      \href{}{\hspace{1mm}Niels ~Leadholm}\footnotemark[1],
      \href{}{\hspace{1mm}Viviane ~Clay}\footnotemark[1],
      \href{}{\hspace{1mm}Scott ~Knudstrup},
      \href{}{\hspace{1mm}Hojae ~Lee},
      \href{}{\hspace{1mm}Jeff ~Hawkins} \\
      Thousand Brains Project, Redwood City, CA, United States \\
      \texttt{\{nleadholm, vclay, sknudstrup, hlee, jhawkins\}@thousandbrains.org}
    }
\fi

\begin{document}

\footnotetext[1]{Joint first authors.}

\maketitle
\begin{abstract}

Current AI systems achieve impressive performance on many tasks, yet they lack core attributes of biological intelligence, including rapid, continual learning, representations grounded in sensorimotor interactions, and structured knowledge that enables efficient generalization. Neuroscience theory suggests that mammals evolved flexible intelligence through the replication of a semi-independent, sensorimotor module, a functional unit known as a cortical column. To address the disparity between biological and artificial intelligence, \textit{thousand-brains systems} were proposed as a means of mirroring the architecture of cortical columns and their interactions.

In the current work, we evaluate the unique properties of Monty, the first implementation of a thousand-brains system. We focus on 3D object perception, and in particular, the combined task of object recognition and pose estimation. Utilizing the YCB dataset of household objects, we first assess Monty's use of sensorimotor learning to build structured representations, finding that these enable robust generalization. These representations include an emphasis on classifying objects by their global shape, as well as a natural ability to detect object symmetries. We then explore Monty's use of model-free and model-based policies to enable rapid inference by supporting principled movements. We find that such policies complement Monty's modular architecture, a design that can accommodate communication between modules to further accelerate inference speed via a novel `voting' algorithm. Finally, we examine Monty's use of associative, Hebbian-like binding to enable rapid, continual, and computationally efficient learning, properties that compare favorably to current deep learning architectures. While Monty is still in a nascent stage of development, these findings support thousand-brains systems as a powerful and promising new approach to AI, and reinforce the importance of sensorimotor learning for developing intelligent systems.

\end{abstract}

\keywords{Sensorimotor \and Embodied \and Active Perception \and Reference Frames \and Representation Learning \and Model-Based \and World Models \and 6 Degrees-of-Freedom Pose Estimation \and Object Recognition \and Object-Centric Representations}

\section{Introduction}
\label{sec1:intro}

Artificial intelligence (AI) has progressed rapidly in the last decade, driven primarily by advances in deep learning architectures and computational scale. Despite significant progress in domains such as image classification, language processing, and game playing, developing a system that exhibits human-like cognitive abilities remains a fundamental challenge. In particular, current systems lack core attributes of biological intelligence, such as rapid learning from limited data \parencite{Lake2015Human-levelInduction}, continuous adaptation to new situations \parencite{McCloskey1989CatastrophicProblem, Flesch2018ComparingMachines}, and robust generalization using structured representations \parencite{Szegedy2014IntriguingNetworks, Geirhos2021PartialVision, schneider2024surprising, gavrikovcan, motamed2025generative}.

A consequence of these limitations is the fundamental difference in how humans and current AI systems learn about the world. While biological systems acquire knowledge through active exploration and continuous sensorimotor integration \parencite{held1963movement, gibson1966senses, Yarbus1967EyeObjects, gilchrist1997saccades}, leading AI architectures rely on passive processing of internet-scale datasets. For example, humans rapidly learn the 3D structure of objects through the active movement of sensors such as their eyes or a finger \parencite{gibson1966senses, Kroliczak2003TheObjects}; only later are such representations used to ground more abstract knowledge such as language \parencite{bruner1974toward, howell2005model, frank2023bridging}. In contrast, state-of-the-art deep learning architectures are trained passively on internet-scale data, where language forms the primary basis for representation learning \parencite{Radford2019gpt2, brown2020gpt3, OpenAI2023gpt4}. After training on orders of magnitude more language data than any human child experiences \parencite{frank2023bridging}, current methods attempt to adapt these architectures to sensorimotor tasks \parencite{Driess2023, Black2024, bjorck2025gr00t}. Despite recent advances, human-like sensorimotor capabilities remain out of reach.

Building algorithms derived from biological intelligence represents an alternative path forward. Vernon Mountcastle proposed that the basis for mammalian intelligence is the replication of a core computational unit, the cortical column \parencite{edelman1982mindful, Mountcastle1997TheNeocortex}. This proposal followed the observation that neighboring neurons in the cortex display functional organization, such as common inputs and extensive, local connectivity \parencite{mountcastle1957, hubel1974uniformity, Mountcastle1997TheNeocortex}. Parallel work demonstrated that cortical columns throughout the brain project to motor regions \parencite{ThalamusBookShermanGuillery2013, usrey2019, Prasad2020LayerTargets}, suggesting sensorimotor loops as a central motif in every region of cortex. Finally, structured representations have long been recognized as core to higher intelligence \parencite{Tolman1948CognitiveMen, biederman1987recognition, Lake2016, Sabour2017DynamicCapsules, Whittington2020TheFormation}, yet how such representations are formed throughout the neocortex has remained unclear.

Building on decades of neuroscience work, the Thousand Brains Theory (TBT) \parencite{Hawkins2019ANeocortex, hawkins2021thousand, Hawkins2025} proposed that each cortical column is a semi-independent sensorimotor system, and that a column learns structured models of the world through movement. As sensors move over objects in the world, information is laid down within a \textit{reference frame}, an explicit coordinate system within which the relative arrangement of sensed information is represented. Through movement of sensory organs such as an eye or a finger, an individual cortical column learns models of entire objects as 3D structures, despite receiving information from only a small sensory patch at any given moment in time. The interaction of many cortical columns in hierarchical and non-hierarchical arrangements can further enable efficient inference and compositional representations, but an individual cortical column remains a powerful computational unit, even on its own.

Work in biological models of perception \parencite{Lewis2019LocationsCells, Bicanski2019ACells, Leadholm2021GridRecognition, rao2024sensory} and robotics \parencite{Pezzementi2011ObjectGeometry, Browatzki2014ActiveRobot, suresh2024neuralfeels} has hinted at the promise of systems that combine sensorimotor learning with reference frames. However, only recently has a system that encapsulates all the principles of the TBT been developed \parencite{clay2024thousand}, an architecture known as a \textit{thousand-brains system}. The first implementation of such a system was given the moniker Monty, in reference to Mountcastle's column theory, and is now available at \url{https://github.com/thousandbrainsproject/tbp.monty/} (MIT License). The present work is the first quantitative demonstration of Monty's capabilities, representing a significant step in applying sensorimotor systems to the challenging task of 3D object perception. 

Monty represents a fundamentally different approach to AI that places embodied, sensorimotor learning at its core. Central to its architecture is the \textit{learning module}, a computational unit derived from the structure of cortical columns. Monty introduces several key technical innovations in the form of:
\begin{itemize}
    \item A primary role for sensorimotor interaction in both learning and inference. Through movement, even very simple sensory inputs can enable learning complex objects, as well as subsequent inference.
    \item The use of an explicit reference frame to build structured, 3D models of objects that can be leveraged for rapid and robust inference. These representations enable generalization by emphasizing the structural form (shape) of objects, and naturally identify symmetries in the world.
    \item The ability to combine model-free policies with policies informed by internal models within each learning module (model-based policies), affording rapid recognition of objects through principled movements.
    \item A communication protocol, referred to as the Cortical Messaging Protocol (CMP), to enable scaling of Monty systems through the addition of modular components. Scaling can accommodate additional learning modules and sensory inputs, affording more rapid inference when combined with a novel consensus-forming algorithm.
    \item The use of Hebbian-like, associative binding for rapid and computationally efficient learning. Updates are sparse and local with respect to internal models, enabling continual learning.
\end{itemize}

These principles make thousand-brains systems uniquely different from existing AI approaches, representing a new way to build intelligent systems.

The following experiments utilize the YCB dataset of household objects \parencite{YCB} to demonstrate the above capabilities. All code for replicating our experiments is available at \url{https://github.com/thousandbrainsproject/tbp.tbs_sensorimotor_intelligence}.

\section{Background and Related Works}
\label{sec2:background}

The challenge of creating intelligent systems that can learn about and recognize objects through sensorimotor interaction spans multiple research areas in machine learning, robotics, and biological perception. Here we review key areas and their relationship to our work.

\subsection{Deep Learning}
Recent years have seen the development of large-scale deep learning systems, particularly in the form of large language and vision models. In the sensorimotor domain, leading approaches leverage internet-scale pretraining to bootstrap the perceptual capabilities of systems such as robots \parencite{Driess2023, Black2024, bjorck2025gr00t}. Notably, the underlying representations are initially learned on passive datasets due to the enormous data requirements of deep learning. In particular, while deep learning algorithms are able to approximate highly complex functions given sufficient training \parencite{jumper2021highly}, they show difficulties in generalizing to out-of-distribution data \parencite{mayilvahanan2024search}. This inability to generalize may relate to their limited use of structured representations. For example, deep learning systems have a bias towards recognizing objects based on texture rather than shape \parencite{Szegedy2014IntriguingNetworks, Geirhos2021PartialVision, gavrikovcan}, and do not robustly develop object-centric representations \parencite{Locatello2020Object-CentricAttention, zimmermann2023sensitivity}, idiosyncrasies that contrast sharply with human perception \parencite{spelke1990principles, Geirhos2021PartialVision}. As such, strong performance in deep learning systems is predicated on densely sampling the data distribution, yet the diverse nature of the real world makes such an approach infeasible as a means of developing sensorimotor intelligence. In contrast, our work demonstrates the value of an inductive bias for structured models, including the ability to generalize to novel poses of objects given little training data, and an innate tolerance to noise.

In addition to requiring large quantities of data for learning, deep learning systems face a related challenge in the setting of continual learning. In particular, deep learning typically assumes that data is sampled from an independent and identically distributed (i.i.d.) dataset. Such an assumption ensures that stochastically sampled inputs provide a reasonable proxy of the true gradient during back-propagation of errors ("back-prop") \parencite{rumelhart1986learning, bottou2010large}, and that they are representative of a static dataset that will not differ in the future. However, real-world conditions typically violate this assumption, such as a stream of inputs where observations are temporally correlated, or changes in the underlying statistics of the world. Such distributional shifts are even more likely when an agent is able to change its behavior, and therefore how the world is sampled. Under such conditions, deep learning systems progressively overwrite their representations in a process known as catastrophic forgetting \parencite{McCloskey1989CatastrophicProblem}. This contrasts with life-long learning in humans, and it is perhaps telling that efforts to explain back-prop as biologically plausible are inconsistent with several facts of neurobiology \parencite{whittington2019theories}. Instead, the established mechanisms for learning in the brain are based on developing associative (Hebbian) connectivity between co-active neurons \parencite{hebb1949, Markram1997RegulationEPSPs, Song2000CompetitivePlasticity, chklovskii2004cortical}. Such learning relies on locally available information, and results in sparse, rather than global, updates to learned connections. Consistent with this, we demonstrate that simple associative learning can support both rapid and continual learning in a sensorimotor system.

Beyond learning mechanisms and internal representations, Monty differs significantly from deep learning approaches in its architecture. For example, a core tenet of deep learning is that a deep hierarchy is necessary for useful representations to emerge \parencite{Lecun2015DeepLearning}. Furthermore, motor control in deep learning systems is typically delegated to a single, monolithic network, separate from sensory processing \parencite{raad2024scaling, Black2024, bjorck2025gr00t}. In contrast, a significant proportion of processing in the cortex relies on a remarkably shallow hierarchy \parencite{suzuki_2023, Hawkins2025}, with motor projections found in seemingly all cortical regions \parencite{ThalamusBookShermanGuillery2013, usrey2019, Prasad2020LayerTargets}. We demonstrate that a shallow sensorimotor architecture can learn generalizable representations, as well as leverage synergistic model-free and model-based policies to guide the actions of Monty.

Finally, deep neural networks for vision tasks typically assume high-dimensional, high-resolution inputs from a large visual area \parencite{Krizhevsky2012ImageNetNetworks, Dosovitskiy2021AnScale}. In contrast, the human fovea provides high-acuity information for only a fraction of the visual field - approximately the size of a fingernail held at arm's length \parencite{o1991thumb, tuten2021foveal}. While leveraging a large, high-resolution input might seem desirable, like the i.i.d. assumption, it is incompatible with a world where all information cannot be simultaneously perceived. In Monty, like in the human cortex, learning operates with a narrow receptive field. Rather than serving as a disadvantage, Monty combines this constrained input with movement to develop structured representations of objects.

\subsection{Reinforcement Learning}
Reinforcement learning (RL) represents another major paradigm in learning, one where the loop of action and perception takes a central role \parencite{sutton1998reinforcement}. However, deep reinforcement learning has traditionally relied on \textit{model-free} approaches, a setting where an action policy is learned without developing an explicit, structured model of the world. While powerful given sufficient training data \parencite{Mnih2015Human-levelLearning}, such representations have shown limited ability to generalize to out-of-distribution tasks and environments. The use of explicit models in RL (\textit{model-based} RL) is a promising approach \parencite{silver2018general, hafner2025mastering}, but learning and leveraging such models represents a significant challenge \parencite{schneider2024surprising}. Our work explores the utility of reference frames for rapidly learning structured representations of objects, before leveraging these to inform a model-based policy that enables rapid inference. We demonstrate the use of such a model-based policy alongside an input-driven, model-free policy.

Note that our approach is not fundamentally at odds with reinforcement learning, and Monty could also leverage this paradigm. Indeed, there is ample neurobiological evidence for reinforcement learning in the brain \parencite{schultz1997neural, sutton2018reinforcement}. However, our focus is on learning general-purpose models, even in the absence of rewards, which policies can then leverage. We leave investigating the synergistic effect of reinforcement learning with Monty's representations to future research.

\subsection{Traditional Robotics} By its very nature, robotics overlaps the paradigms of active perception and sensorimotor learning \parencite{bajcsy2018revisiting}. Presented with real-world challenges such as limited data quantities and non-i.i.d. distributions, structured representations are also frequently leveraged in robotics, including mapping algorithms such as Simultaneous Localization and Mapping (SLAM) \parencite{thrun2008simultaneous}. As such, while Monty is designed for general perception and representation learning \parencite{clay2024thousand}, our present work most closely relates to prior research in robotics.

In particular, the task setting we consider of identifying a perceived object as well as its pose can be formulated as what "environment" an embodied agent is observing, and from where. In other words, object recognition and pose detection can be viewed as a \textit{localization} problem, an approach proposed in both \textcite{Pezzementi2011ObjectGeometry} and \textcite{Browatzki2014ActiveRobot}. This formulation enables leveraging methods such as particle filter localization (also known as Monte Carlo localization) \parencite{thrun2001robust, thrun2008simultaneous} for object inference. The proposal that a single cortical column in the brain (and therefore a learning module in Monty) builds reference frames of objects, and recognizes them by localizing a position within the reference frame \parencite{Hawkins2019ANeocortex, Lewis2019LocationsCells, clay2024thousand}, therefore relates to this important work. A key distinction is that the TBT proposes that \textit{all} objects, from those held in a hand, to abstract concepts of society and mathematics, are represented with such reference frames.

In the task domain of object recognition and pose estimation (the scope of the present work), \textcite{Browatzki2014ActiveRobot} is most similar due to their focus on recognizing 3D objects with a particle filter system. At the same time, there are many fundamental differences between the approach we adopt, and this prior work. For example, the authors were concerned with localization on the 2D surface of a view sphere around an object, rather than a 3D location on an object, as we consider. As such, their input features were large, 2D key-frame images of an object. In contrast, we use 3D poses extracted from a narrow receptive field as the primary input feature. This important design choice forces Monty to learn with movement and locally observable information, ultimately emphasizing the 3D structure of objects. Additionally, while \parencite{Browatzki2014ActiveRobot} recognized that localization could be used to predict the pose of an object, they did not evaluate the performance of their algorithm in this respect, nor did they explore related concepts such as symmetry. Finally, their work leveraged a single monolithic system, without an ability to add more sensory inputs where available. In contrast, Monty implements a modular system together with a novel "voting" algorithm to enable accommodating additional sensory inputs as desired \parencite{clay2024thousand}.

More recently, \textcite{suresh2024neuralfeels} explored the role of combining locally sensed tactile information with SLAM for 3D object perception. However, this work was concerned with the online learning of 3D objects for the purpose of reconstructing their shape in a given interaction episode, rather than retaining models for later recognition of the objects. Indeed, even within this single-episode paradigm, retaining key-frames and periodically retraining on these was required to avoid catastrophic forgetting in the system due to the use of a deep learning architecture. Finally, the use of local tactile information was considered an adjunct to a global view of the object, rather than the primary method of perception as we consider here.

\subsection{Models of Biological Perception}

Biological perception is inherently sensorimotor; for example, kittens deprived of an active role in movement do not develop normal vision \parencite{held1963movement}. \textcite{Hawkins2019ANeocortex} proposed the Thousand Brains Theory (TBT) of the neocortex, arguing that individual cortical columns use neurons similar to grid cells \parencite{Hafting2005MicrostructureCortex} to build structured models of objects. Biological models of perception have since demonstrated the ability for reference frames to support learning and recognition, including for abstract, synthetic objects \parencite{Lewis2019LocationsCells}, as well as in the setting of simple 2D datasets such as MNIST \parencite{Bicanski2019ACells, Leadholm2021GridRecognition, rao2024sensory}. In contrast to this prior work, our results do not focus on biological realism at an implementation level, but on demonstrating robust recognition on a diverse dataset of complex 3D objects, including pose detection. Monty is capable of such tasks due to a variety of technical developments. These include methods for tracking thousands of hypotheses, the transformation of sensed features by inferred object poses, and the use of a model-based policy for principled movement, among others. As such, Monty represents an important demonstration that algorithms inspired by theories of cortical function can perform challenging tasks in a 3D world, revealing competitive advantages over non-biological approaches.

\section{Methods}
\label{sec:methods}

\subsection{Monty Architecture Overview}
\label{sec:monty_architecture}

For a detailed description of Monty and its underlying algorithms, we refer the reader to \textcite{clay2024thousand}, as well as the associated code-base (\url{https://github.com/thousandbrainsproject/tbp.monty}) and documentation (\url{https://thousandbrainsproject.readme.io}). Below, we provide a high-level overview of the architecture, as well as those details that are most relevant to the results presented here. 

Monty is a modular, sensorimotor learning system consisting of three primary components: learning modules (LMs), sensor modules (SMs), and a motor system. These components interact through a standardized communication protocol called the Cortical Messaging Protocol (CMP) \parencite{clay2024thousand}. The purpose of these components can be summarized as follows:

\begin{itemize}
    \item Sensor Modules: Process raw sensory input into CMP-compliant input data.
    \item Learning Modules: Build object models, and use these to infer the identity and pose of objects in the world. Internal to each learning module is a Goal State Generator (GSG) that can output CMP goal states to influence the motor system. If there are multiple LMs, they can modulate one another's hypotheses via CMP votes.
    \item Motor System: Generates actions based on its current state and any received goal states. While the motor system receives CMP-based goal states, it outputs actions as actuator-specific motor commands.
    \item CMP Message: Standardized message consisting of a pose and features. This pose is provided in a common reference frame. The above components only use CMP-compliant messages to communicate with one another. The use of the CMP enables straightforward modification of Monty, such as the inclusion of additional sensor modules and learning modules.
\end{itemize}

\begin{figure}[htbp]
    \centering
    \includegraphics[width=0.55\textwidth]{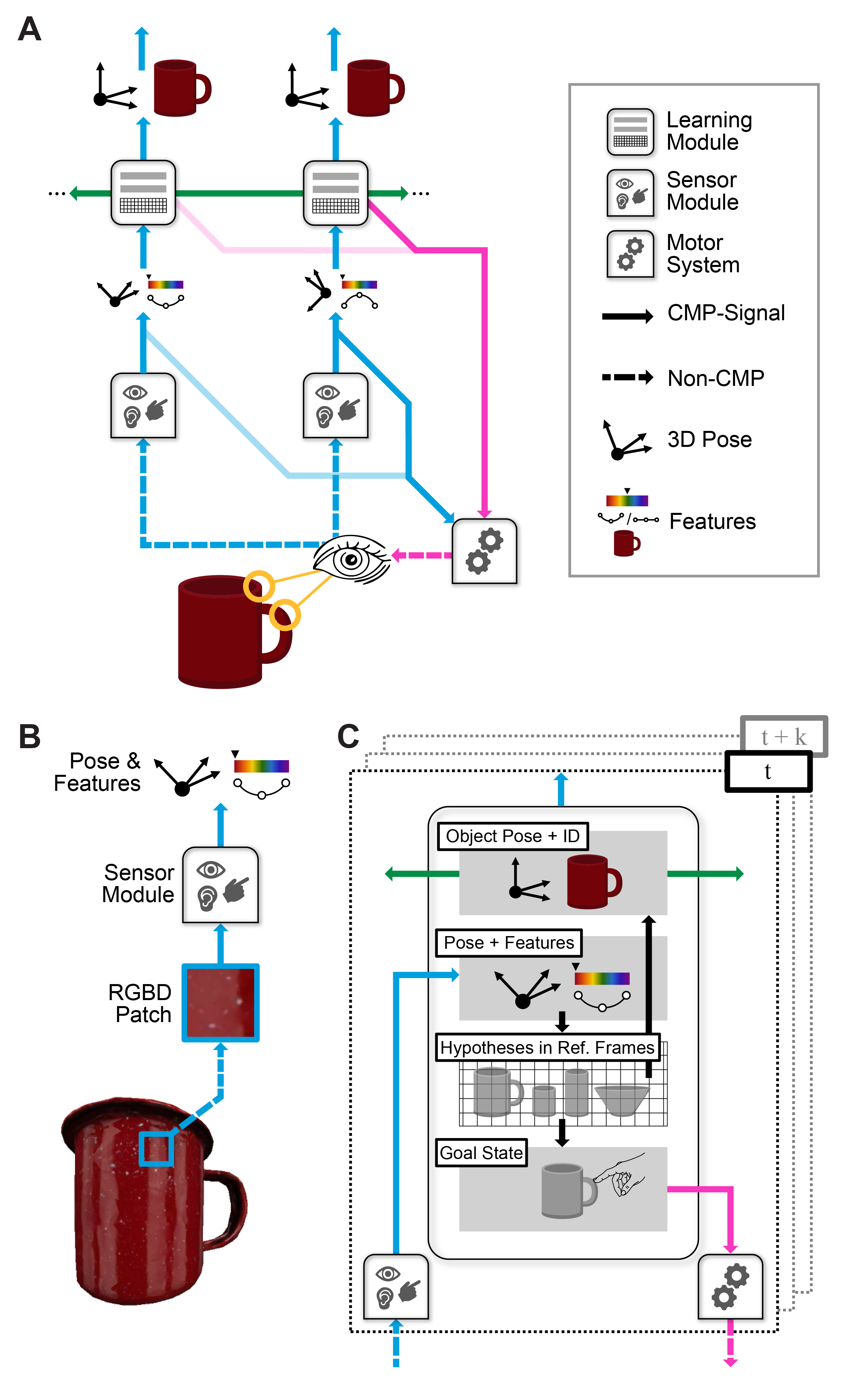}
    \caption{\textbf{The Architecture of Monty, a Thousand-Brains System.} A) Monty consists of a series of modular components, the learning module, sensor module, and motor system, that interact via a common Cortical Messaging Protocol (CMP). CMP signals consist of a pose and features, and can be used for feed-forward communication (blue), lateral voting for achieving rapid consensus (green), and goal states for guiding the motor system (pink). The pose in CMP signals is communicated in a common reference frame (e.g., body-centric). In addition to goal states from LMs, SMs can send sensory data to the motor system to support model-free policies. B) In the example of vision, a sensor module receives non-CMP, RGBD data from a narrow receptive field (patch), before converting this into a CMP-compliant format consisting of a pose (location + orientation of the local surface patch) and features at that pose. C) Detailed overview of a learning module, shown at a time-point $t$. The incoming pose and features are used to establish hypotheses in the reference frames of known objects. Movement (derived from the current and previous incoming pose) is used to update the hypothesized location in the reference frame, before matching the latest pose + feature information against learned representations. Any given hypothesis in the reference frames is associated with an evidence value, which is adjusted depending on whether incoming movement and sensory data are consistent with the stored representations. Information about current hypotheses can be used to infer the pose and ID of the observed object, and to vote on this with other LMs. Internal models can also be used to generate goal states for how the agent should move in the world. Inference is therefore a sensorimotor process where the current action output will influence the next sensory input.}
    \label{fig:overview_diagram}
\end{figure}

Figure \ref{fig:overview_diagram}A illustrates the high-level architecture of Monty and how these components interact. 

Below, we provide a more detailed, mathematical overview of Monty. A summary of the mathematical notation we use is provided in the Appendix in Tables \ref{tab:math_notation} and \ref{tab:math_notation_2}.

\subsection{Learning}

Learning and inference operate under similar principles in Monty. During a given \textit{episode}, Monty interacts with an object over a cycle of perception and movement. This interaction can be for the purpose of learning about an unknown object, or to recognize a previously encountered one, but we begin by describing the former setting. We will also consider learning given a Monty system with a single learning module (LM), although the principles are similar when multiple LMs are learning at the same time.

In the following, we will refer to multiple coordinate systems. For clarity, $B$ indicates a shared, body-centric coordinate system, $M$ indicates the coordinate system of an object model, and $S$ indicates the coordinate system defined by local features on a surface patch. We use the standard adopted in \textcite{craig2009introduction}, where the prescripts in $\prescript{B}{S}{\mathbf{R}}$ indicate the orientation of coordinate system $S$ relative to $B$.

At each episode step $t$, a sensor module (SM) will be located somewhere in the environment, and will pass its perceived information to its associated LM. When doing so, the SM processes raw sensory input into a CMP message $\phi_t$. Each CMP message contains a pose defined in a shared, body-centric coordinate system $B$, alongside optional, non-pose features. SMs are thus tailored to transform a particular sensed modality into a CMP-compliant message.

In the case of Red-Green-Blue-Depth (RGB-D) data and the present work, this output consists of an observed rotation given by the surface normal and directions of principal curvature \parencite{porteous2001geometric}, which together define a locally observed rotation, $\prescript{B}{S}{\mathbf{R}}_{t} \in SO(3)$. This is provided together with a location ($\prescript{B}{}{x_t} \in \mathbb{R}^3$), which is also defined in the body-centric coordinate system $B$. Additional features that can be included in a CMP signal are non-pose information ($n_t$) such as hue, saturation, and value (HSV), or the magnitudes of principal curvature. Note that this information is provided for a single point at the center of the SM's receptive field. It is thus both low-dimensional, and is derived from a narrow region of space (e.g., $64 \times 64$ pixels from a zoomed-in view, see Figure \ref{fig:overview_diagram}B). The full CMP message is given by:

\begin{equation}
    \phi_t = \{\prescript{B}{}{x_t}, \prescript{B}{S}{\mathbf{R}}_{t}, n_t\}
\end{equation}

During an episode, an object is presented at a particular rotation in the environment. Let us define $m$ as the object identity and $\prescript{B}{M}{\mathbf{R}} \in SO(3)$ as the rotation of the object's coordinate system $M^m$ in the shared coordinate system $B$. For clarity, the $m$ superscript will generally be omitted from $M$ when $M$ appears in super or subscript. In supervised learning, Monty is provided with the ground-truth values of the object ID and orientation. To learn a new object model, Monty moves over the object, enabling the LM to associate CMP observations $\phi_t$ with locations in an internal reference frame, informed by the ground-truth object ID and pose.

More concretely, when an object has not yet been learned, the LM initializes a new reference frame for the object. This consists of a 3D Cartesian coordinate space where sensed features ($\prescript{B}{S}{\mathbf{R}}_{t}$ and $n_t$) are associated with an internal location that is currently active ($\prescript{M}{}{x_t} \in \mathbb{R}^3$). Following learning, the model of an object $m$ will then be defined by the set of three-tuple points as follows:

\begin{equation}
    \label{eq:learning_binding}
    \mathcal{M}^m = \left\{ (\prescript{M}{}{x_i}, \prescript{M}{S}{\mathbf{R}_i}, n_i) \right\}
\end{equation}

Each point in the model (indexed by $i$) is therefore associated with a location ($\prescript{M}{}{x_i}$), a locally observed pose ($\prescript{M}{S}{\mathbf{R}_i}$), and a set of features ($n_i$) found at that location. Both the location and pose (e.g., the pose of a surface normal and the principal curvature directions) are in the reference frame of the object. For a new point learned at step $t$ and indexed by $i$, $\prescript{M}{}{x_i} \coloneqq \prescript{M}{}{x_t}$, $\prescript{M}{S}{\mathbf{R}}_{i} \coloneqq (\prescript{B}{M}{\mathbf{R}})^{-1} \prescript{B}{S}{\mathbf{R}}_{t} = \prescript{M}{B}{\mathbf{R}} \prescript{B}{S}{\mathbf{R}}_{t}$ and $n_i \coloneqq n_t$. We note that $\prescript{B}{S}{\mathbf{R}}_{t}$ is transformed by the rotation of the object to align with the internal reference frame, but the pose of the object as a whole ($\prescript{B}{M}{\mathbf{R}}$) is not stored for each point.

To update the active location $\prescript{M}{}{x_t}$ during learning, the LM derives a movement vector ($\prescript{B}{}{v_t}$) from the two most recently received CMP signals, $\phi_t$ and $\phi_{t-1}$, given by $\prescript{B}{}{v_t} = \prescript{B}{}{x_t} - \prescript{B}{}{x_{t-1}}$. To align this displacement with the internal model, it is transformed by the provided rotation of the object, giving $\prescript{M}{}{v_t} = (\prescript{B}{M}{\mathbf{R}})^{-1}\prescript{B}{}{v_t}$. The internal location $\prescript{M}{}{x_{t-1}}$ is then updated by integrating the movement vector ($\prescript{M}{}{x_t} = \prescript{M}{}{x_{t-1}} + \prescript{M}{}{v_t}$), i.e., by performing \textit{path integration}, also known as dead reckoning \parencite{whishaw2001dead, Hafting2005MicrostructureCortex}. As the location space is an internal coordinate system,  the choice of $\prescript{M}{}{x_{t=0}}$ during learning is arbitrary.

By the above method, Monty's LM develops a structured representation of an object's features, and their relative arrangement, through movement. The key step in this process is the binding between active sensory features and active location representation in equation \ref{eq:learning_binding}, an instantaneous operation analogous to associative or conjunctive binding, and the simplest possible form of Hebbian learning.

\begin{figure}[htbp]
    \centering
    \includegraphics[width=0.8\textwidth]{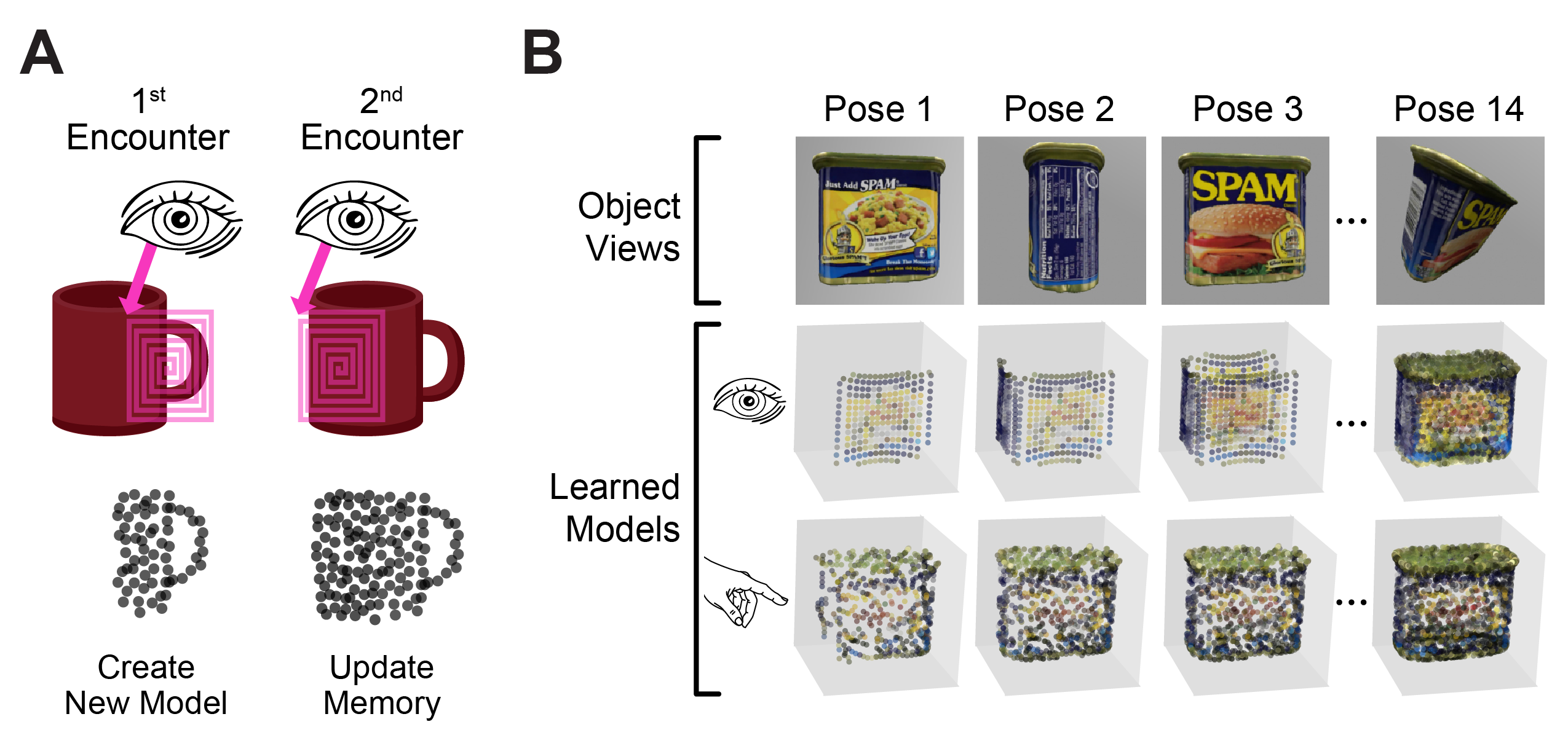}
    \caption{\textbf{Learning in Monty.} A) Monty performs supervised learning by initializing an internal coordinate space (reference frame) for an object if it has not been previously learned. It then moves over the object, associating sensed observations with locations in this internal reference frame. The movement pattern used during learning consists of either an eye-like agent performing a spiral scan (as shown diagrammatically) or the movement of a finger-like agent following the object's surface and its contours. If the object has been encountered before, observations are transformed so as to align with the previously established reference frame, enabling the model to be extended with new information. B) In our experiments utilizing the YCB objects, Monty views each object under 14 rotations, corresponding to the six faces and eight corners of a cube. We show the models learned by Monty when coupled with a `distant', eye-like agent (top row), and those learned when coupled with a `surface', finger-like agent (bottom row). Although in both instances Monty has access to RGB-D data, the choice impacts Monty's exploration strategies during learning and inference, discussed in further detail in Section \ref{sec:movement}.}
    \label{fig:monty_learning}
\end{figure}

In any given episode, Monty is given a finite number of steps to explore the surface, and depending on the policy used during learning, it may not observe all parts of the object. As such, each object is presented over multiple episodes, where each episode corresponds to a different rotation of the object. If Monty has already observed an object, it uses the episode's ground truth rotation ($\prescript{B}{M}{\mathbf{R}}^\text{gt}$) to transform the newly learned observations into the previously initialized reference frame for object $m$ (Figure \ref{fig:monty_learning}A). In the majority of our experiments, Monty views objects in 14 different rotations during learning (Figure \ref{fig:monty_learning}B), corresponding to the six faces and eight corners of a cube. Naturally, Monty may observe highly similar locations on an object more than once, either due to its exploration policy or redundant views. To prevent dense object models, we further add the constraint that a given three tuple $(\prescript{B}{}{x_t}, \prescript{B}{S}{\mathbf{R}_t}, n_t)$ only forms a new point $(\prescript{M}{}{x_i}, \prescript{M}{S}{\mathbf{R}_i}, n_i) \in \mathcal{M}^m$ if it is sufficiently displaced in 3D Cartesian or feature space from existing points in $\mathcal{M}^m$.

Finally, we note that Monty can perform unsupervised learning when object ID and pose labels are omitted. In such a setting, learning is proceeded by a sensorimotor inference phase to determine what object and pose is present \parencite{clay2024thousand}. We have described Monty's learning setup for object pose and ID recognition with supervised learning, and episodes and epochs are defined accordingly. Other applications of Monty may be aimed at interaction with the world, where inferring objects and poses is only a sub-problem, and rigid definitions of epochs and episodes would be relaxed. However, unsupervised learning, as well as such evaluations, lies beyond the scope of the present work.

\subsection{Inference}

Inference operates under a similar principle to learning - Monty moves over an object during an episode, with sensory information passed from SMs to LMs. To perform inference, LMs use learned object models to accumulate evidence for different \textit{hypotheses} about the identity and orientation of the object currently being observed. Hypotheses also consider the location in the object's reference frame that the learning module is observing. More concretely, each LM $l$ maintains a set of $K$ hypotheses at every step $t$ defined as:

\begin{equation}
    \mathcal{H}_t^l = \{(m_k, \prescript{B}{M}{\mathbf{R}_k}, \prescript{M}{}{x_{k,t}}, e_{k,t})\}_{k = 1}^K
\end{equation}

where $k$ signifies the hypothesis index, $m_k$ is the hypothesized object identity, $\prescript{B}{M}{\mathbf{R}_k}$ is the hypothesized rotation of the object in the shared coordinate system, $\prescript{M}{}{x_{k,t}}$ is the hypothesized location within the reference frame of object $m$, and $e_{k,t} \in \mathbb{R}$ is the evidence score for hypothesis $k$. Note that both the location and evidence score associated with a hypothesis change as a function of the episode step $t$. The evidence scores thereby serve as a proxy for a non-parametric distribution over the possible objects, their rotations, and the locations within their reference frames, similar to a particle filter \parencite{thrun2001robust, thrun2008simultaneous}.

As during learning, each learning module derives a movement vector $\prescript{B}{}{v_t}$ from the two most recently received CMP signals. For each hypothesis $k$, this movement is transformed by the hypothesized rotation of the object $\prescript{B}{M}{\mathbf{R}_k}$ to align with the object's internal reference frame (i.e., $\prescript{M}{}{v_{k,t}} = \prescript{B}{M}{\mathbf{R}_k}^{-1} \prescript{B}{}{v_t}$), providing the updated location $\prescript{M}{}{x_{k,t}} = \prescript{M}{}{x_{k,t-1}} + \prescript{M}{}{v_{k,t}}$. In this way, Monty is able to move over an object while maintaining hypotheses about its current location in the object's reference frame.

Following the integration of movement, the LM compares the received CMP message (observation) to any stored information near the updated location. The evidence score $e_{k,t}$ for each hypothesis is then adjusted based on how well the current observation matches what the object model predicts at that specific location. The result is an increase in the score when there is a good match (supporting the hypothesis) or a decrease when there is a poor match (contradicting the hypothesis).

In particular, let $\mathcal{X}^m = \{\prescript{M}{}{x_1}, \prescript{M}{}{x_2}, ... \prescript{M}{}{x_N}\}$ be the set of learned points for object $m$. We begin by identifying all such points within an $\varepsilon$ neighborhood of the hypothesized location, that is:

\begin{equation}
    \mathcal{N}_\varepsilon(\prescript{M}{}{x_{k,t}};\mathcal{X}^m) = \{ \prescript{M}{}{x_i} \mid \|\prescript{M}{}{x_i} - \prescript{M}{}{x_{k,t}}\| < \varepsilon \}
\end{equation}

For each hypothesis, we will modify the evidence value as a function of the distance in pose information between the locally observed $\prescript{B}{S}{\mathbf{R}}_{t}$ and the stored $\prescript{M}{S}{\mathbf{R}_i}$. To compare these, $\prescript{B}{S}{\mathbf{R}}_{t}$ is transformed by the hypothesized rotation of the object to align it with the learned local pose. We also update the evidence value according to the distance between observed ($n_t$) and stored ($n_i$) non-pose features, however, only a mismatch in pose features can cause evidence values to \textit{decrease}. More concretely:

\begin{equation} 
\label{eq:delta_R}
    \Delta e^\mathbf{R}_{k,t} \propto D(\prescript{B}{M}{\mathbf{R}_k}^{-1}\prescript{B}{S}{\mathbf{R}_t}, \prescript{M}{S}{\mathbf{R}_i})
\end{equation}

and

\begin{equation}
\label{eq:delta_n}
    \Delta e^n_{k,t} \propto D(n_t, n_i) \text{ where } \Delta e^n_{k,t} \in \mathbb{R}_{\geq 0}
\end{equation}

where $D$ are distance functions, and $\Delta e$ is the change in evidence value, which can occur as a function of distance in either pose ($\mathbf{R}$) or non-pose ($n$) features (i.e., $\Delta e_{k,t} = \Delta e^\mathbf{R}_{k,t} + \Delta e^n_{k,t}$). Thus, when matching observations to learned models, the orientation of the observation is privileged over non-pose information, as its contribution can be negative, causing hypotheses to be gradually eliminated. This reflects the emphasis in Monty (modeled on human behavior \parencite{Geirhos2021PartialVision}) that the spatial structure of objects is more important than non-spatial details. For example, a heart symbol is defined by its structural form (shape). Even if the color red is most commonly associated with it, it can easily be recognized in any other color. In other words, the ID is causally defined by the shape, while color is merely correlated with it. To correctly identify objects based on their shape, classification therefore emphasizes surface-composing pose features, consistent with human biases \parencite{Wagemans2012AOrganization, lonnqvist2025contour}.

As a final detail of inference, we note that SMs can be parameterized to only pass information to their LM when sensed features have changed significantly, subject to feature-specific thresholds. This ensures that LMs only receive and process information when it is likely to convey new information that changes their hypotheses.

Before we discuss how Monty moves in Section \ref{sec:movement}, we now turn to initialization and convergence.

\subsubsection{Initialization}
\label{sec:initialization}

At the start of inference, the hypothesis space $\mathcal{H}^l_{t=0}$ for learning module $l$ is initialized based on alignment between the observed input $\phi_0$ and stored features, without the use of movement. In particular, given that the hypothesis space uses internal coordinates, movement is not meaningful until hypothesized locations in the reference frame exist. All initial locations are derived from the learned points in an object's reference frame, that is $\prescript{M}{}{x_{k,t=0}} \in \mathcal{X}^m$, although following later movement, hypothesized locations will rarely align exactly with points in $\mathcal{X}^m$.

For each location hypothesis $\prescript{M}{}{x_{k,t=0}}$, initial observations help define the pose hypotheses, as a valid object pose would transform the observed local pose such that it matches the stored local pose. This enables inferring poses directly, rather than sampling from a set of previously experienced or predefined object poses. More concretely, we define $\prescript{B}{M}{\mathbf{R}_k}$ such that:

\begin{equation}
    \prescript{B}{S}{\mathbf{R}_{t=0}} = \prescript{B}{M}{\mathbf{R}_k}\prescript{M}{S}{\mathbf{R}_i} \\
\end{equation}

giving

\begin{align}
\label{eq:derive_rotation_hypothesis}
    \prescript{B}{M}{\mathbf{R}_k} &= \prescript{B}{S}{\mathbf{R}_{t=0}} (\prescript{M}{S}{\mathbf{R}_i})^{-1} \\
    &= \prescript{B}{S}{\mathbf{R}_{t=0}} (\prescript{M}{S}{\mathbf{R}_i})^{T}
\end{align}

following the property that the inverse of a rotation matrix is its transpose. Through this process, locations in the learned reference frame are associated with hypotheses about the object rotation ($\{\prescript{B}{M}{\mathbf{R}_k}, \prescript{M}{}{x_{k,t=0}}\} \in \mathcal{H}^l_{t=0}$).

When initializing points, we note that a given local surface pose ($\prescript{B}{S}{\mathbf{R}_{t=0}}$) is ambiguous, either due to the symmetry of principal curvature directions, or where principal curvature directions are undefined (e.g., on a flat surface, what is known as an umbilical point \parencite{porteous2001geometric}). As such, a point $\prescript{M}{}{x_i}$ will result in multiple location hypotheses, each with a different object rotation that satisfies equation \ref{eq:derive_rotation_hypothesis} subject to the assumptions made about $\prescript{B}{S}{\mathbf{R}_{t=0}}$.

We emphasize again that through the above process, rotation hypotheses $\prescript{B}{M}{\mathbf{R}_k}$ are not sampled from a fixed, learned distribution (e.g., from ([0, 0, 0], [45, 0, 0], ..., [315, 315, 315])), but are instead inferred directly. As we will demonstrate, this enables the recognition of poses that have never been encountered during learning, without using a prohibitively large search space.

\subsubsection{Convergence}
\label{sec:convergence}

After initialization, inference consists of a series of movements and sensory inputs. At any given time, Monty will have a most likely hypothesis (MLH), indicated by hypothesis index $k^* = \arg\max_{k \in K} e_{k,t}$. Following a period of steps, the evidence associated with the MLH may surpass all other evidence values by a variable threshold $\theta_\text{converge}$. That is:

\begin{equation}
\label{eq:convergence}
    e_{k^*,t} - \theta_\text{converge} > e_{j,t}, \forall j \in \{1, \dots, K\}, j \ne k^*
\end{equation}

If this occurs, the LM reaches a terminal condition at episode step $t$, outputting the MLH as the pose and ID for comparison to the ground-truth. An inference episode may also terminate due to sufficient elapsed steps (e.g., $t=500$) before the threshold condition is reached, in which case the MLH at the final step is used to evaluate Monty's performance.

Note that the terminal condition can be parameterized to emphasize efficient inference, or high accuracy. For example, should a situation require high accuracy with fewer constraints on computational costs, $\theta_\text{converge}$ can be increased. Also note that since Monty has a most likely hypothesis at every step, a classification and corresponding confidence can be extracted at any time during an episode. In this work we do not tune $\theta_\text{converge}$ for different evaluations, instead aiming for an overall balance between accuracy and computational efficiency.

We also define a relative threshold $\theta_\text{update}$ that determines which hypotheses are updated at each step. If a hypothesis falls below this threshold, relative to the evidence count of the MLH, then we save computational resources by not updating this hypothesis. Once again, we use the same parameter value for all of our experiments reported here.

An additional feature of Monty is the ability to naturally detect rotational symmetry. In particular, we define a set of rotation hypotheses:

\begin{equation}
    \mathcal{R}^{m} = \left\{ \prescript{B}{M}{\mathbf{R}}_k \mid e_{k,t} \geq e_{k^*,t} - \theta_\text{converge} \right\}_{k=1}^K
\end{equation}

That is, this is the set of $|\mathcal{R}^{m}| = J$ rotations associated with high evidence hypotheses, and the system will therefore not reach its normal terminal condition (equation \ref{eq:convergence}) while $J > 1$. We define the number of consecutive steps during which there is minimal change in this set as $\tau_{\text{sym}}$. If this counter surpasses a threshold ($\theta_\text{sym}$), and all hypotheses are for the same object $m$, Monty will output that these poses are symmetric and terminate the episode.

This leads to a natural definition of symmetry as \textit{poses that cannot be distinguished by sensorimotor exploration of an object}, or what we term sensorimotor symmetric (SMS). When an object is deemed SMS, we measure Monty's rotation error as the difference between the ground truth rotation ($\prescript{B}{M}{\mathbf{R}^\text{gt}}$) and the minimally distant rotation in $\mathcal{R}^{\text{obj}}$. That is:

\begin{equation}
    E^{\text{rot}} = \min_{j \in \{1, \dots, J\}} D_\text{geo}\left( \prescript{B}{M}{\mathbf{R}}_j, \prescript{B}{M}{\mathbf{R}^\text{gt}} \right)
\end{equation}

where $D_\text{geo}$ is the geodesic distance, i.e., the relative angle between the two rotation matrices. A trivial solution for low rotation errors would be to set $\theta_{\text{sym}}$ arbitrarily low, after which symmetry will be declared with $|\mathcal{R}^{m}| \gg 1$, i.e., many rotation hypotheses will be in the set when $\tau_{\text{sym}} > \theta_{\text{sym}}$. However, we will show that for a reasonable value of $\theta_{\text{sym}}$, this does not occur and Monty does indeed detect interpretable symmetry while avoiding false positives.

\subsection{Movement and Policies}
\label{sec:movement}

Movement is central to how Monty learns about and understands the world. The motor system in Monty generates and executes actions $a_t$ that enable interaction with the environment. Actions can consist of primitives including a rotation, translation, or (in simulated environments) moving directly to a location in absolute coordinates.

The motor system's actions are carried out by an \textit{agent}, which we define as a set of sensors associated with an actuator. In a human, an agent would include an eye or the tip of a finger, where each of these sensory organs is associated with a set of sensory patches that move together. In Monty, these biological structures are approximated with a \textit{distant} agent, and a \textit{surface} agent. Like a camera rotating on a point or an eye performing saccades, the distant agent observes objects by pivoting its sensors. On the other hand, the surface agent observes objects from a shorter distance, and uses sensed surface normals to orient itself towards the surface before moving tangentially. Through such movements, it continuously follows the surface of objects. Unlike a biological finger, the surface agent can also perceive color.

In the current experiments, Monty is associated with only one agent at any given time, although it is designed with multiple agents in mind \parencite{clay2024thousand}. Furthermore, the majority of experiments make use of the distant agent, unless noted otherwise.

\subsubsection{Model Free Policies}

During learning, the distant agent follows a scanning, spiral-like policy to densely sample observations of an object at a given rotation. At inference, its orienting movements cause it to sample observations in the form of a random walk on the visible (non-occluded) portion of an object's surface. The only use of sensory input is to reverse the previous action when a move off of the object is detected, which is determined by a sudden change in depth values.

The surface agent follows the same policy during learning and inference. In addition to orienting towards surface normals, this agent's motor system implements an innate, model-free policy for following areas of prominent curvature. In particular, this curvature-guided policy uses sensed principal curvatures to follow prominent features such as the rim or handle of a mug, similar to heuristics observed in humans \parencite{gibson1966senses}.

\subsubsection{Model Based Policies}
\label{sec:model_based_method}

Whether the motor system is coupled with the surface agent or distant agent, it is able to receive CMP-compliant \textit{goal states} from an LM's goal-state generator (GSG) in order to enact principled movements. In particular, a goal state defines a desired state (orientation and location) for an agent to occupy. The GSG is the component of an LM that uses the LM's learned models to propose goal states.

In the present work, LMs implement a model-based policy called the \textit{hypothesis-testing policy}. During inference, an LM's GSG can use the most likely hypotheses to propose a goal state that would disambiguate them. Recall that the MLH is the single hypothesis defined as:

\begin{equation}
    k^* = \arg\max_{k \in K} e_{k,t}
\end{equation}

Within LM $l$, this MLH will be associated with a given object $m$ and hypothesized rotation $\prescript{B}{M^m}{\mathbf{R}}_{k^* \in \mathcal{H}^{l,m}_t}$. The first way the hypothesis-testing policy can be leveraged is to disambiguate the MLH from the second most likely object with ID $q \neq m$ and rotation hypothesis $\prescript{B}{M^q}{\mathbf{R}}_{h^* \in \mathcal{H}^{l,q}_t}$, where $h^*$ is the MLH index for object $q$. To do so, the learned points for the two objects, $\mathcal{X}^m$ and $\mathcal{X}^q$ are aligned using the respective most likely locations, then transformed by the respective most likely rotations. This has the effect of enabling points in each model to be compared in a shared, body-centric space. The objects are positioned in this shared space such that the origin corresponds to the locations in each object where Monty considers itself most likely to be. That is, the points are updated such that:

\begin{equation}
   \prescript{B}{}{\mathcal{X}^m} = \left\{\prescript{B}{}{x_{m,i}} \mid \prescript{B}{}{x_{m,i}} = \prescript{B}{M^m}{\mathbf{R}}_{k^*} (\prescript{M^m}{}{x_i} - \prescript{M^m}{}{x_{k^*})} \right\}_{i=1}^{|\mathcal{X}^m|}
\end{equation}

and

\begin{equation}
   \prescript{B}{}{\mathcal{X}^q} = \left\{\prescript{B}{}{x_{q,j}} \mid \prescript{B}{}{x_{q,j}} = \prescript{B}{M^q}{\mathbf{R}}_{h^*} (\prescript{M^q}{}{x_j} - \prescript{M^q}{}{x_{h^*})} \right\}_{j=1}^{|\mathcal{X}^q|}
\end{equation}

Note the additional indexing of the points $\prescript{B}{}{x}$ by object ID $m$ and $q$, given they are now in a shared coordinate space.

The GSG then determines the point in the MLH object model that maximizes the distance between itself and its nearest neighbor in the second model. This is the point whose observation is most likely to disambiguate the hypotheses. The index of this point $\prescript{M^m}{}{x_i}$ is given by:

\begin{equation}
    i^* = \operatorname*{argmax}_{i} \left\{ \min_{j} \left\| \prescript{B}{}{x_{m,i}} - \prescript{B}{}{x_{q,j}} \right\| \right\}
\end{equation}

Intuitively, if the MLH is a mug and the second most likely object a soup can, the policy will select a point $\prescript{M^m}{}{x_{i^*}}$ on the handle of the mug. The nearest neighbor of this point in the transformed $\prescript{B}{}{\mathcal{X}^q}$ would be on the outer wall of the can. A similar process can be used by the GSG to test a hypothesis that distinguishes the two most likely poses for the single most likely object. 

Following the above, the policy specifies a goal state that would result in the sensor observing point $\prescript{M^m}{}{x_{i^*}}$, assuming the hypothesized rotation is correct. This goal state is then passed to the motor system, whose responsibility it is to carry out action primitives that result in achieving the goal state. As we evaluate performance in simulation, the motor system executes a motor primitive that moves it directly in absolute coordinates to this goal state (a `jump'), before normal movement is resumed. 

If the hypothesis-testing policy is available, the GSG will generate goal states subject to certain conditions being satisfied. These include that the LM has performed a minimum number of steps, and a sufficiently high evidence count for the MLH. 

If there are multiple LMs, any LM's GSG can output a goal state, a form of distributed motor planning. To coordinate these, goal states are associated with a confidence value that represents how important it is to act on the goal state. As a simple heuristic, this confidence is the evidence value of an LM's MLH, normalized by the number of observations the LM has received. If the motor system receives multiple goal states, it selects the one with the highest confidence, increasing the likelihood that its action is informed by a reasonable estimate of the world's state.

This policy can be viewed as performing the most discriminative action, similar to variance-maximizing approaches used in prior work on localization \parencite{fairfield2008active, Browatzki2014ActiveRobot}, albeit here in a setting of 3D objects with ambiguous rotations, and with multiple learning systems coordinating their goal states.

\subsection{Voting}
\label{sec:voting_methods}

While Monty can perform all learning and inference using only a single LM, multiple LMs can work together to enable faster inference. In particular, if there are multiple LMs in a Monty system, they share evidence for their hypotheses through lateral connections in a process known as \textit{voting} \parencite{Hawkins2017, clay2024thousand}. This enables LMs to quickly reach consensus based on having observed different parts of an object.

Let us define a set of $L$ learning modules. For the experiments we consider in this work, multi-LM systems have all-to-all connectivity, and connections are therefore bidirectional.

We denote the LM sending votes as $\hat{l}$. For two LMs, we will then consider the votes sent by LM $\hat{l}$ to LM $l+1$. These votes are derived from hypotheses $\mathcal{H}^{\hat{l}}_t = \{(m_k, \prescript{B}{M}{\mathbf{R}}_k, \prescript{M}{}{x_{k,t}}, e_{k,t})\}_{k = 1}^K$. However, to be useful to the receiving LM, they must be transformed.

To do so, the displacement between the sensor modules that connect to each LM is used to update the location hypotheses ($\prescript{M}{}{x_{k,t}}$) associated with outgoing votes. More concretely, LMs $\hat{l}$ and $l+1$ will receive observations from the SMs containing locations $\prescript{B}{}{x_t}^{\hat{l}}$ and $\prescript{B}{}{x_t}^{l+1}$, where once again we note that these locations are in a shared, body-centric coordinate system. We can therefore define a displacement, but rather than occurring across time (i.e., a movement $\prescript{B}{}v_t = \prescript{B}{}{x_t} - \prescript{B}{}{x_{t-1}}$), this is the instantaneous separation of the sensory observations, defined as $\prescript{B}{}{d_t} = \prescript{B}{}{x_t}^{l+1} - \prescript{B}{}{x_t}^{\hat{l}}$. This displacement is then transformed by the hypothesized object rotation in LM $\hat{l}$, providing $\prescript{M}{}{d}_{k,t} = (\prescript{B}{M}{\mathbf{R}}_k)^{-1} \prescript{B}{}{d_t}$. Finally, this transformed displacement is used to update the location for voting, given by $\prescript{M}{}{\hat{x}_{k,t}} = \prescript{M}{}{x_{k,t}} + \prescript{M}{}{d}_{k,t}$.

In addition to the above transformation, the LM normalizes the evidence counts $\{e_{k,t} \mid k = 1...K\}$ to the range [-1.0, 1.0] before sending out votes.

When the votes $\mathcal{H}^{\hat{l}}_t$ are received by LM $l+1$, it determines whether the locations of its internal hypotheses align with the locations provided by incoming votes. Evidence values are incremented proportional to the nearness of these locations. In practice, $\hat{l}$ will only send a subset of the $K$ possible hypotheses for each object, sub-selecting those with higher evidence counts subject to a threshold.

Note that the result of LM $l$ sharing a transformed location  $\prescript{M}{}{\hat{x}_{k,t}}$ is that voting does not operate as a simple bag-of-features for whether two LMs are sensing the same object. For example, it is not sufficient that both LMs observe random parts of a mug for the their hypotheses of a mug to grow stronger. Instead, the incoming votes must agree with LM $l+1$'s hypotheses about \textit{where} the object is in the world, which in turn is affected by the potential rotation of the object. As such, voting operates in a similar manner to Monty's structured accumulation of evidence during sensorimotor perception, while reducing the actual need for movement.

Despite (indeed because of) this sensitivity to spatial structure, the relative arrangement of SM-LM pairs do not need to be fixed for voting to operate. For example, the LMs associated with two digits, one for each hand, could still vote with one another, with $\prescript{B}{}{d_t}$ updated depending on the current positions of each finger.

We highlight that a vote can be considered as a set of CMP signals, where the object ID and evidence values are the non-pose features. As such, voting is agnostic as to the sensory modality that underlies the representations in two laterally connected LMs. This theoretically enables voting across sensory modalities \parencite{clay2024thousand}, although we leave demonstrating this capability to future work.

\subsection{Training and Evaluation Setting}

For training and evaluation, we present isolated instances of the YCB objects \parencite{YCB}, a dataset of 77 common household objects. Monty's interactions with the objects are mediated in the Habitat simulator \parencite{savva2019habitat, szot2021habitat, puig2023habitat}. Sensor modules receive data from simulated RGB-D cameras with a resolution of $64 \times 64$ pixels. These images correspond to zoomed-in, narrow patches on the object surface. In addition, a wider `view-finder' image is available to help initialize the agent position at the start of an episode, or to move back onto the object if Monty loses contact with it. However, sensory information in the view-finder is not provided to any LMs for the purpose of inference or learning.

\subsection{Other Methods}

\subsubsection{Vision Transformer Networks}
\label{sec:vit_methods}

The final part of our work, Section \ref{sec:rapid_learning}, includes comparisons to deep learning architectures, specifically Vision Transformers (ViTs) \parencite{Vaswani2017AttentionNeed, Dosovitskiy2021AnScale}.

For the majority of our results, we use the ViT model \parencite{Dosovitskiy2021AnScale} pre-trained on ImageNet-21k (14 million RGB images at resolution $224 \times 224$, 21k+ classes) \parencite{deng2009imagenet}. To adapt the model to the RGB-D setting, we create a new encoding channel for depth, initialized using the mean weights for the RGB channels. To enable pose prediction alongside object classification, we replace the classification head to accommodate a multi-objective loss function, providing object labels and ground-truth rotations during learning. The objective function is given by:

\begin{equation}
L = L_{\text{cls}} + \lambda L_{\text{rot}}
\end{equation}

where $L_{\text{cls}}$ represents cross-entropy loss, $\lambda$ represents a weighting factor, and $L_{\text{rot}}$ represents the geodesic loss between the predicted and ground truth unit quaternions, in keeping with prior work \parencite{xiang2017posecnn}.

All other weights, including positional encodings, remain unaltered at initialization from the base models, however during fine-tuning, we enable all weights to be updated. For some of our experiments, we vary the model size from the smaller \texttt{ViT-b32-224-in21k} to the much larger `ViT huge'(\texttt{ViT-h14-224-in21k}); unless noted otherwise, we use \texttt{ViT-b16-224-in21k}, which achieved the best balance between accuracy and model size.

Training and inference data consists of $224 \times 224$ RGB-D images of the YCB objects extracted from Habitat, such that the full object is in view and with a black background. Unless noted otherwise, the training data consists of 14 images for each object, where each image is captured with the object at one of the rotations also used by Monty during learning. We split the dataset comprising 14 images $\times$ 77 objects into training and validation sets with an $80:20$ ratio when optimizing hyperparameters for improved performance. Before evaluation, we train on the full dataset of $14 \times 77$ samples, as for Monty. We do not use data augmentation methods in order to match the datasets received by Monty and the ViTs, the same reason for which we constrain the ability of Monty to move freely during learning and inference. In evaluation, we used a set of 5 novel rotations for each object, again corresponding to the same rotations used to evaluate Monty. For both Monty and the ViT models, we only calculate the rotation error at inference where the predicted class is correct.

Using the validation dataset, we performed extensive hyperparameter tuning to optimize the performance of the ViT. This included establishing an early stopping point of 25 epochs of training, as well as an optimal learning rate of 5e-4. We also identified improved performance through the use of the AdamW \parencite{loshchilov2017decoupled} rather than the Adam optimizer \parencite{Kingma2015Adam:Optimization}, gradient clipping of 1.0, and the inclusion of a layer-norm \parencite{ba2016layer} in the final classification head. We use a learning rate scheduler with a warm-up phase followed by cosine decay, although the network's performance was less sensitive to this choice. For architectural variants, we also tried training with the primary backbone frozen, and appending a multi-layer output head, but these did not improve performance. Finally, we experimented with varying $\lambda$ (the coefficient balancing the loss for classification vs. pose prediction); adjusting the parameter in either direction resulted in a drop in performance on the other task, and so we opted for a middle-ground value of 1.0.

Our results also include a model trained from scratch, where we follow the same architecture and hyperparameters as above, but begin with randomly initialized weights throughout the network. For this model we re-established optimal values for early stopping and the learning rate on the validation subset, identifying 75 epochs for early stopping, and a learning rate of 1e-5.

Finally, for our continual learning experiments, we leverage the pre-trained transformer. For the first task, where accuracy is already 100\% owing to masking of the softmax, we train for an arbitrarily chosen 10 epochs, although in practice there are no gradients available for learning at this point, and the model weights do not change. For the other tasks, we perform early stopping when training accuracy achieves 90\% on the current task. This enables the model to achieve reasonable accuracy on the current task, while minimizing catastrophic forgetting, which is aggravated by attempting to achieve higher accuracy.

\subsubsection{Estimating Computational Efficiency}
\label{sec:flops_methods}

To quantify the computational efficiency of models, we estimated the number of floating point operations (FLOPs) required during inference and learning. These estimates exclude any overhead associated with acquiring data samples, in particular the FLOPs involved in running the Habitat simulator.

For Monty, we implemented a custom Python library able to track FLOPs by a combination of operation interception and function wrapping, with a focus on Numpy, the source of the vast majority of Monty's operations. We augment this with custom handling for operations such as k-d tree construction and querying which are not otherwise captured by this approach. Details are available at our public repository (\url{https://github.com/thousandbrainsproject/tbp.floppy}), including how we estimate the FLOPs associated with numerical operations such as trigonometric functions.

For the ViT models, we estimate inference FLOPs using the Pytorch package \texttt{calflops} \parencite{calflops}. To estimate FLOPs for training, we extrapolate FLOPs following the method of \textcite{kaplan2020scaling}. In particular, we begin by measuring the FLOPs associated with inference (i.e., the forward pass) on a single $224 \times 224$ RGB image using \texttt{calflops}. Following \textcite{kaplan2020scaling}, we estimate that training FLOPs for a single image is $3\times$ inference FLOPs, and then extrapolate based on the total number of images in the dataset, together with the number of epochs of training. In the case of pretraining on ImageNet-21k, this corresponds to 14 million images \parencite{deng2009imagenet} and 90 epochs \parencite{Dosovitskiy2021AnScale}.

\subsubsection{Additional Hyperparameters}

For additional details on hyperparameters for Monty and the ViT models, or to replicate our experiments, we direct the reader to the repository available at \url{https://github.com/thousandbrainsproject/tbp.tbs_sensorimotor_intelligence}.

\section{Results and Discussion}

\subsection{Robust Inference}

We begin by demonstrating that Monty is able to leverage its properties as a sensorimotor system to perform inference under a variety of adverse conditions.

\subsubsection{Sensorimotor Inference}

\begin{figure}[htbp]
    \centering
    \includegraphics[width=0.8\textwidth]{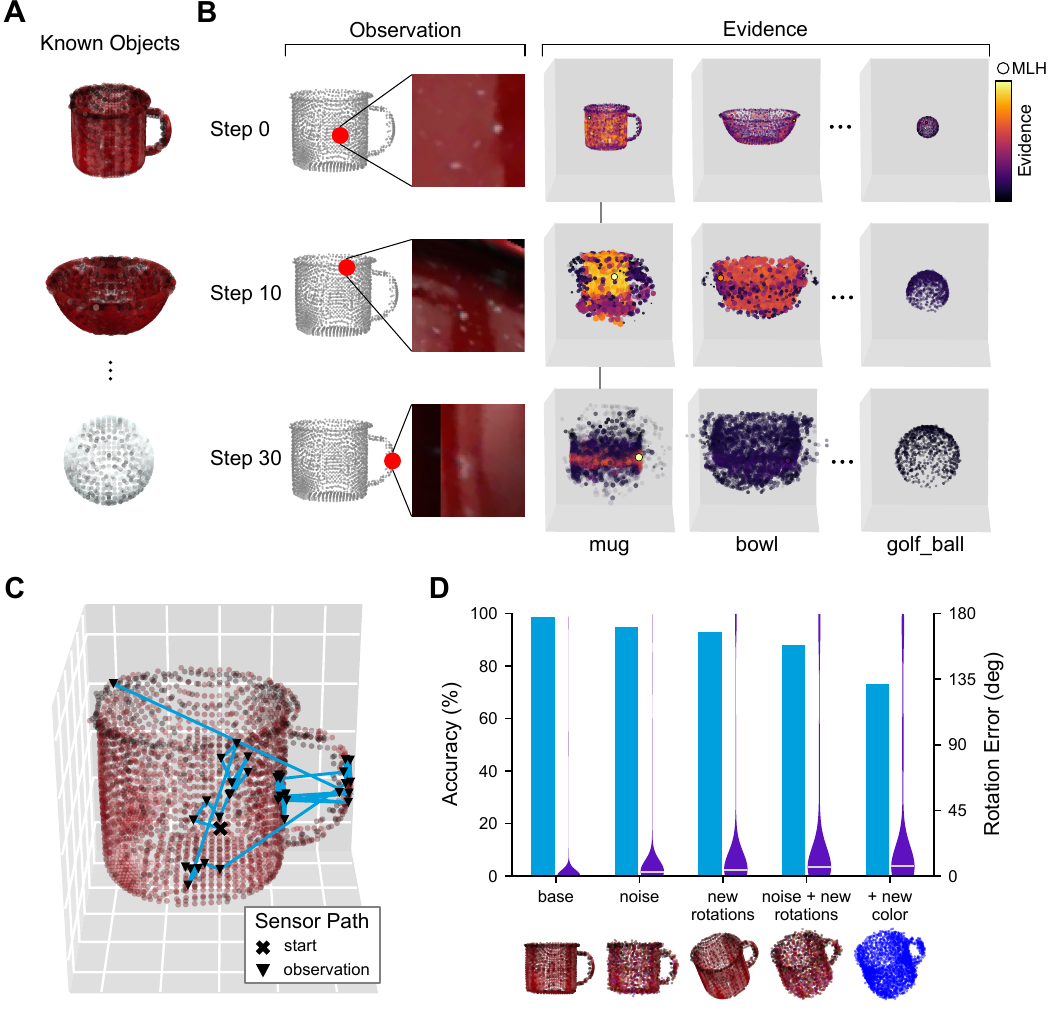}
    \caption{\textbf{Robust Sensorimotor Inference.} A) We examine Monty's internal representations during inference of a mug, focusing on the evidence values associated with three out of the 77 known objects. B) Incoming sensory information from local patches arrives in the form of a CMP signal. An LM matches this input against information stored in its reference frames for known objects. Evidence accumulates for different location hypotheses on an object, conditioned on hypothesized rotations of the object. When initialized, these evidence values are relatively uniform before integrating subsequent movements and sensations. However, the most likely hypothesis (MLH) quickly converges to an accurate representation of the agent's location on the object. C) Monty recognizes objects by performing a series of movements coupled with sensory observations. Shown is the path that realized the observations shown in (B), where Monty first moves onto the handle of a mug, then makes a large movement onto the rim, before moving back to the handle. D) Monty is able to recognize all 77 YCB objects with high accuracy (98.6\%), including predicting their rotation in space. Introducing perturbations in the form of feature noise, new rotations, or a combination of these has minimal impact on performance. Recognition is still successful for most objects (73.1\%) even when noise and new rotations are compounded by entirely changing textural information (`new color'). This adverse condition is achieved through a uniform blue HSV value for all observations on all objects. None of these perturbations were experienced during learning, and they are thus all out-of-distribution with respect to training. Gray bars indicate the median.}
    \label{fig:robust_inference}
\end{figure}

We have argued that a sensorimotor system coupled with internal reference frames should develop structured representations that enable robust generalization. To demonstrate this, we first visualize an example of Monty's internal representations during an episode of inference. Figure \ref{fig:robust_inference}A-B  shows the evidence associated with object locations as Monty explores a mug in Habitat, highlighting the evidence values for three objects that are known to Monty. Following a series of movements and sensory observations, evidence values for locations on incongruent objects quickly fall, leaving only locations on the mug with a high evidence count (Figure \ref{fig:robust_inference}B). Importantly, the relative movements between sensations are crucial to recognizing the object. For example, local features on the mug (red patches of curved and flat surfaces) are non-specific and would be consistent with the bowl if viewed as an unordered list, i.e., a bag-of-features. However, the combination of features with their relative arrangement is unique to the true object, a property that Monty leverages.

Figure \ref{fig:robust_inference}C shows a sample path taken by Monty during inference. By moving over the object, rather than passively observing whatever limited sensory information is first received, Monty can efficiently eliminate ambiguity caused by object self-occlusion, as well as move to local features that disambiguate an object's ID and pose. In this example, both the rim and handle are useful regions to explore when disambiguating the mug from other cylindrical objects such as the soup can. By performing path integration, Monty can explore objects via novel paths that were not experienced during learning, while still maintaining an accurate estimate of its location on the object. Monty is therefore not constrained to following a fixed (e.g., raster scan) series of movements over an object.

\subsubsection{Rotation Invariance, Noise Robustness, and Generalization}

To evaluate Monty's robustness, we conduct experiments where we assess its ability to classify the 77 YCB objects. In the first condition, we present the objects at rotations observed during learning, and without any additional noise introduced. These experiments utilize an instance of Monty that moves using the distant agent action space and the hypothesis-testing policy. Note that even though the objects are presented in the learned orientations, the sequence of actions used to explore them is different, and hence, Monty will experience a new set of points on the object. We measure i) whether Monty's MLH matches the correct class label, and ii) how closely the rotation output by Monty aligns with the ground-truth rotation (see Methods Section \ref{sec:methods} for details). In Figure \ref{fig:robust_inference}D, we can see that Monty easily performs this baseline task, with a classification accuracy of 98.6\%, and a median rotation error of 0 degrees.

We then introduce adversarial conditions in the form of noise and novel rotations. For the former, we perturbed a variety of critical feature inputs, with noise sampled from a Gaussian distribution or binary symmetric channel (BSC) model \parencite{mackay2003information}. The parameters for these perturbations are provided in the Appendix in Table \ref{tab:robustness_config}. As an intuitive example, we highlight the noise applied to sensed locations. Monty relies on the location feature ($\prescript{B}{}{x_t}$) to integrate movements between subsequent observations. We add Gaussian noise with 2mm standard deviation to all observed locations, in addition to the other noise perturbations described in Table \ref{tab:robustness_config}. Figure \ref{fig:robust_inference}D demonstrates minimal impact on classification accuracy or pose detection when these are introduced (95.1\% classification accuracy, median rotation error 3 degrees). We highlight that Monty is not trained with any form of noise exposure. Instead, robustness appears to emerge from the use of structured representations when performing inference, as the noise introduced does not disrupt the global shape of the objects.

To evaluate robustness to novel rotations, we present each object in 14 random rotations for each YCB object, where each rotation is sampled uniformly from $SO(3)$. In Figure \ref{fig:robust_inference}D, we see that the accuracy of classification and pose prediction remains largely unaffected (93.0\% classification accuracy, median rotation error 4.5 degrees). To understand why, we note that Monty initializes its rotation hypotheses conditioned on sensory observations (described in Methods Section \ref{sec:initialization}). As such, it can accurately predict poses that were never experienced during training. In later results, we will see further examples of how Monty's approach to inferring pose enables generalization beyond the training distribution. Additionally, we consider the condition of combined feature noise and novel rotations (Figure \ref{fig:robust_inference}D), a setting where Monty still achieves strong performance (88.1\% classification accuracy, median rotation error 6 degrees). 

To further measure Monty's reliance on global structure over visual textures for recognition, we include a setting with a uniform HSV value for all observations. In this condition, feature noise (e.g., to the locations) and novel rotations are still included, while all HSV values observed on all objects are set to a uniform value, corresponding to an intense blue color. Note that Monty has never learned the YCB objects in novel colors, and many of these would be difficult even for humans to distinguish without textural information, given similarity in object shapes. Despite this, Monty still achieves high classification accuracy (73.1\% classification accuracy, median rotation error 7 degrees), consistent with a reliance on the global shape of objects when performing recognition.

For all of our following results in the paper, unless noted otherwise, we evaluate Monty using the noise condition, and 5 novel rotations that were selected due to deviating significantly from the 14 canonical views used during training.

\subsubsection{Structured Representations and Object Shape}

\begin{figure}[htbp]
    \centering
    \includegraphics[width=0.8\textwidth]{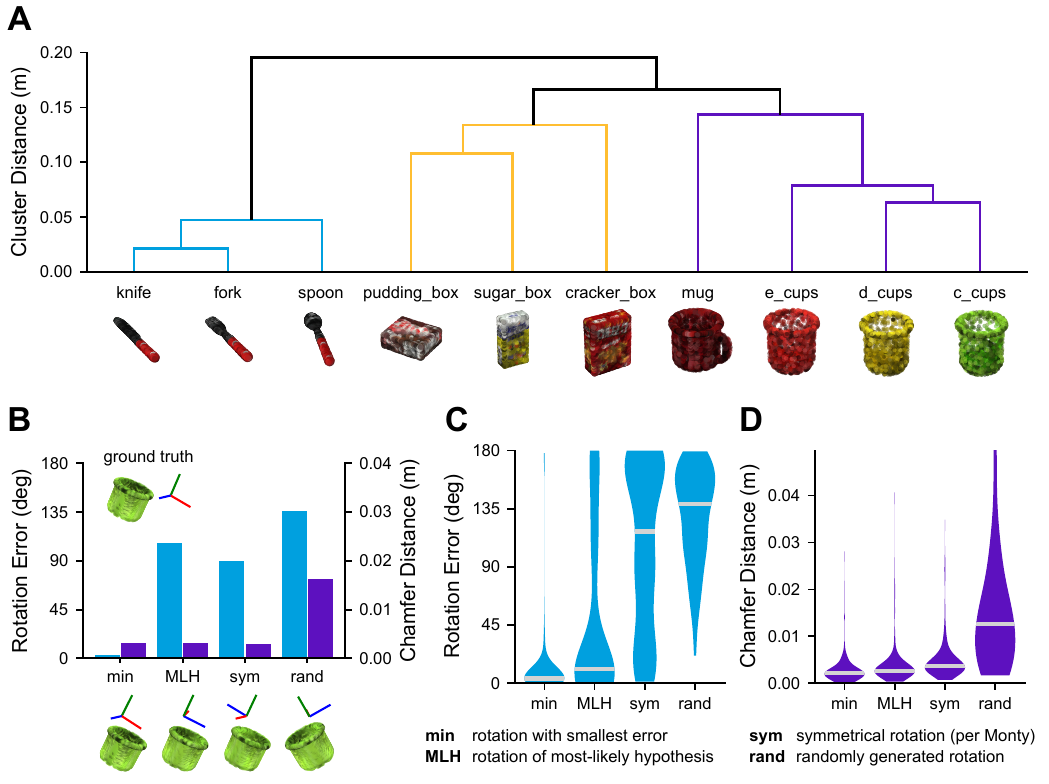}
    \caption{\textbf{Structured Object Representations and Symmetry.} A) We present ten objects to Monty that correspond to human categories of cutlery, boxes, and cups. After Monty explores a given object, we examine the evidence counts for all the hypotheses to measure similarity, where similar evidence counts result in shorter cluster distances. Plotting these in a dendrogram reveals alignment with morphologically meaningful groupings. B) During inference, Monty recognizes when multiple poses are mutually consistent, i.e., the object is sensorimotor symmetric (SMS). Shown here is a single example of three poses deemed SMS by Monty, denoted min, MLH, and sym. Min is the SMS pose with the lowest geodesic distance to the ground-truth rotation. MLH is the SMS pose associated with the most likely hypothesis (highest evidence count). Finally, sym is a third, SMS pose that is neither the MLH nor min rotation. These results show that such SMS poses are indeed symmetric when inspected visually (bottom) or measured by Chamfer distance (model point cloud alignment). When reporting the rotation error in degrees, we therefore use the SMS rotation closest to the ground-truth (min). C+D) Rotation measures derived from a set of inference episodes, where Monty views all 77 objects at 5 novel rotations each, and with feature noise present. These results provide further support that SMS poses correspond to meaningful symmetry in objects, as quantified by the Chamfer distance, even if naive measurements with geodesic distance would appear high.}
    \label{fig:structured_reps_symmetry}
\end{figure}

To explore the basis of Monty's robustness, we further evaluate its representational emphasis on shape. Figure \ref{fig:structured_reps_symmetry}A examines the relationship between different object representations when Monty views 10 selected objects that, to a human, cluster into morphological categories of cutlery, boxes, and cups. As Monty explores an object, it updates evidence counts for all of its hypotheses. When a given object is shown (e.g., the fork), we can examine how similar the evidence count for the fork is to the evidence for other objects, such as the knife or the cracker box. The observed clustering supports the hypothesis that Monty emphasizes the global shape of objects during classification, similar to humans.

We highlight that any given observation provides only a local pose ($\prescript{B}{S}{\mathbf{R}}_t$), defined by the surface normal and principal curvature directions, along with low-dimensional HSV and curvature magnitude values. This information is highly ambiguous when considered without the context of global shape. It is thus notable how the lack of a handle and similar sizes of the c\_, d\_, and e\_cups cause these to cluster with one another more than with the mug. On the other hand, the d\_cup sits far away from the sugar box, despite both being predominantly yellow objects. We emphasize that, given the local ambiguity of observations, Monty's use of movement is crucial for integrating information to develop such globally coherent shape representations.

Such a reliance on shape for classification is present despite training on only 77 objects, presented at 14 rotations each. It is noteworthy that deep learning classifiers trained on orders of magnitude more data demonstrate a consistent bias towards recognizing objects based on texture rather than their shape \parencite{Geirhos2021PartialVision, gavrikovcan}. Monty's use of reference frames and sensorimotor interaction is key to structured representations emerging from such small amounts of training data. In later sections, we will further examine the efficiency with which Monty develops robust representations.

\subsubsection{Representing Symmetry}

Next, we turn to Monty's representation of rotational symmetry. In particular, rotations of certain objects can be ambiguous due to symmetry, a concept familiar to humans, yet challenging to capture in artificial systems \parencite{higgins2022symmetry}. As described in Section \ref{sec:convergence} of our Methods, Monty naturally reports that a set of rotations is symmetric if it is unable to distinguish them through sensorimotor exploration, a condition that we term \textit{sensorimotor symmetric} (SMS). Figure \ref{fig:structured_reps_symmetry}B demonstrates qualitatively that such SMS poses are indeed symmetric when visually inspected. To further quantify this, we use Chamfer distance, a measure that represents the average distance between nearest neighbors in two point clouds. Here we rotate learned models by their predicted rotations, and measure the Chamfer distance between the points in each. Results in Figure \ref{fig:structured_reps_symmetry}B-D provide quantitative support that the point clouds of SMS rotated objects are largely indistinguishable, making it a reasonable indicator of true symmetry. Notably, the development of symmetry representations does not assume internal access to full, ground-truth models of every object during learning. This contrasts with prior attempts to handle symmetry in deep learning systems, where loss functions require access to such ground-truth models \parencite{xiang2017posecnn, wang2019densefusion}.

The finding supports our choice to report rotation error using the SMS pose that minimizes the geodesic distance to the ground-truth rotation. We note that the value of the hyperparameter $\theta_{\text{sym}}$ determines how quickly Monty declares that there is symmetry. As discussed in Section \ref{sec:convergence}, setting $\theta_{\text{sym}}$ to an arbitrarily small value would result in the reported rotation errors always being low. However, this would result in high Chamfer distances (as seen with random rotations in Figure \ref{fig:structured_reps_symmetry}B-D). As insurance against this outcome, we use the same value of $\theta_{\text{sym}}=5$ steps throughout our results.

We highlight this natural handling of symmetry in Monty, as its significance goes beyond establishing low estimates of rotation error. In particular, symmetry has the potential to dramatically improve the efficiency and generalizability of representations in learning systems \parencite{higgins2022symmetry}. For example, when LMs communicate with one another laterally or in a hierarchy, symmetric rotations can be summarized with a single value, avoiding the need to relearn representations following transformations that do not meaningfully change an object's properties. More concretely, one can imagine an agent learning to write with a pencil. Rotating it along its axis of symmetry (rolling it in between one's fingers) has no effect on the position of key components, and therefore the ability of the pencil to write. As such, there is no need for the communicated representation to change; if it did, the agent would be forced to relearn any mappings that support using the object. On the other hand, if the pencil rotates along its long axis (bringing the eraser to the front), then this \textit{does} affect the positioning of its parts, and new learning is appropriate.

Capturing when transformations are symmetric vs. not, as in the pencil example above, relates to the challenging trade-off of representations that are \textit{invariant} vs. \textit{equivariant} \parencite{DiCarlo2007UntanglingRecognition, hinton2012does, Bengio2013RepresentationPerspectives, higgins2022symmetry}. Invariant representations are insensitive to input changes, while equivariant representations \textit{do} change following input changes, and both serve useful properties depending on how they reflect changes in the world. Monty appropriately captures both, showing invariance in its object ID representations as a function of rotation, noise, and visual textures (Figure \ref{fig:robust_inference}B), as well as invariance to symmetric rotations. On the other hand, Monty demonstrates appropriate equivariance to non-symmetric rotations that could prove functionally significant in downstream tasks.

\subsection{Rapid Inference}

We now turn to the efficiency with which Monty can recognize objects, and in particular, the role of its motor policies and the voting algorithm in enabling this.

\subsubsection{Policies for Rapid Inference}

As we have already emphasized, Monty is a sensorimotor system, and movement is crucial to how it learns representations, as well as how it performs inference. However, the significance of how exactly Monty chooses to move is an important aspect which we have not yet explored.

As a naive baseline, Monty can randomly tilt a camera-like sensor located at a fixed base, resulting in perceived locations following a random walk over an object's surface. This is the baseline we consider in the case of the distant agent when it lacks the hypothesis-testing policy. Such a random walk is sufficient to explore the parts of an object that are visible from a given viewing point (i.e., all parts where there is no self-occlusion of the object). However, a random walk is naturally inefficient, potentially revisiting previously observed locations, and taking time to identify useful features. It is also not possible to eliminate self-occlusion, making it difficult to distinguish objects that are ambiguous from a given perspective, such as the mug vs. a cup when the handle is not visible.

\begin{figure}[htbp]
    \centering
    \includegraphics[width=0.8\textwidth]{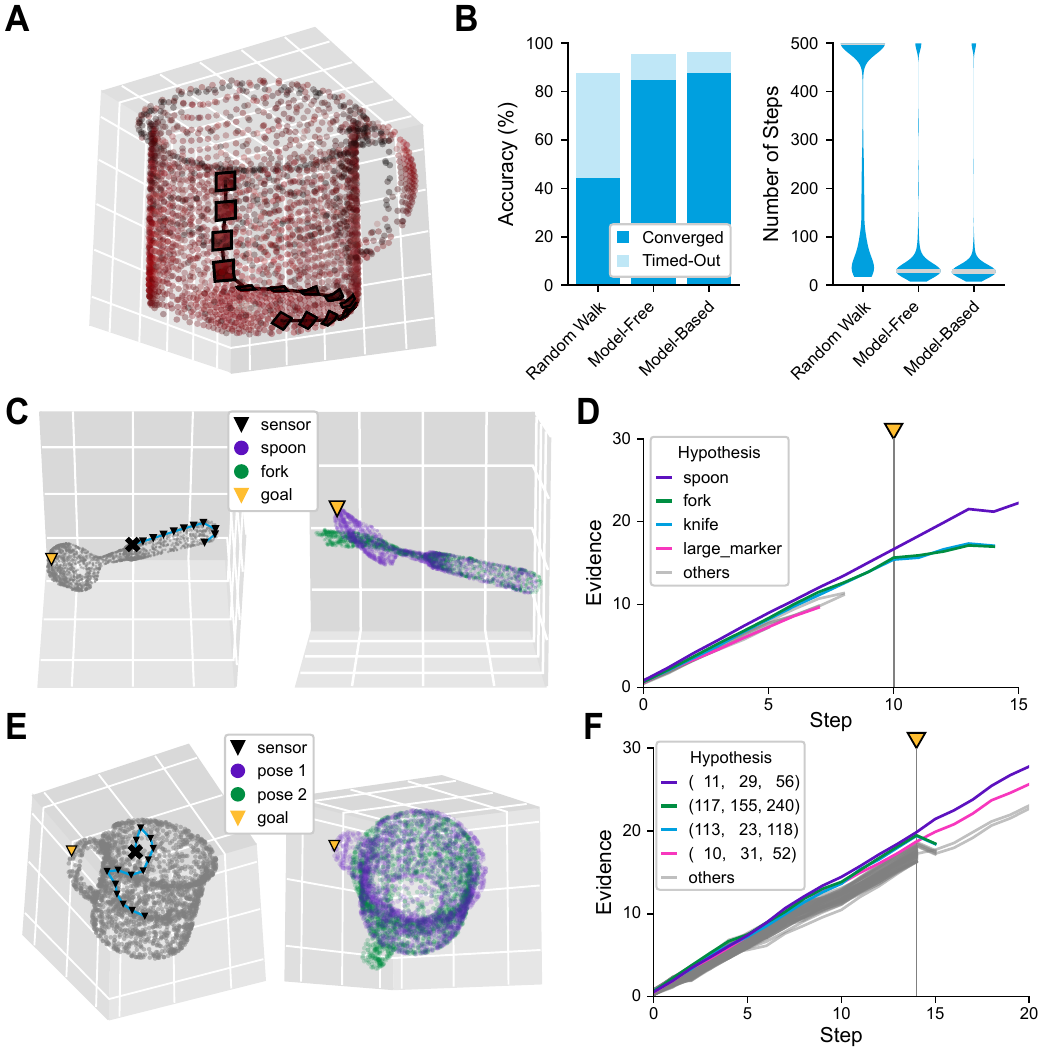}
    \caption{\textbf{Rapid Inference with Model-Free and Model-Based Policies.} A) Rather than perform a random walk over the surface of the object, Monty can make use of model-free policies, resulting in more rapid and robust inference. Here, the agent moves along the surface of the object, similar to a finger; using sensed curvature directions, it is biased to follow directions with significant principal curvature, such as the long axis of a mug's body, as well as its bottom edge. The windowed regions show the input to the sensor module, underscoring how narrow and ambiguous any given sensation is for Monty. B) Using the model-free curvature-following policy and the model-based hypothesis-testing policy boosts Monty's accuracy (left), and reduces the steps required for convergence (right). Note that the most likely hypothesis (MLH) can be correct (i.e., the highest evidence count hypothesis matches the ground-truth object) even if Monty has not reached a terminal condition, although this reflects a state of lower confidence. C) An example of the hypothesis testing policy in action. Monty first senses the handle of a spoon (black path), then uses the hypothesis-testing policy to rapidly disambiguate the spoon from the fork by moving to the goal-state location (gold triangle). D) The evidence counts of Monty's hypotheses, showing the rapid shift when the goal state generated by the hypothesis-testing policy is enacted (vertical line with gold triangle). E-F) The hypothesis testing policy can also disambiguate the poses of a single object, such as two rotations of the mug.}
    \label{fig:rapid_inference_policy}
\end{figure}

\textbf{Model-free policies} are one important means for intelligent systems to select actions. As per their name, such policies do not make use of learned models of the world, yet it is believed that many complex actions that humans perform can be driven by model-free algorithms \parencite{thorndike1911animal, schultz1997neural}. In Monty, the surface agent uses incoming sensory information to both i) orient itself along a surface while moving across it, and ii) follow directions of principal curvature when these are encountered. This enables it to efficiently explore the entire surface of an object, as well as move along regions of interest. As the same policy is used during learning and inference, Monty is also more likely to revisit regions that were encountered during learning, where the object will therefore be better represented in the model. This operates as an innate model-free policy, mirroring heuristics observed in humans when recognizing objects by touch \parencite{gibson1966senses}. Figure \ref{fig:rapid_inference_policy}A shows an example path where the policy first follows the long axis of the mug's side, before following the curvature present on the bottom edges.

Figure \ref{fig:rapid_inference_policy}B demonstrates that introducing this model-free policy already proves very effective. In particular, Monty is able to reach its terminal condition more frequently and in fewer steps. As described in Methods Section \ref{sec:convergence}, Monty reaches a time-out condition after a maximum number of steps (here 500) have taken place. When this occurs, its classification output is still correct if the MLH matches the target object, however this represents a low-confidence outcome. In contrast, Monty can converge before timing out if it becomes sufficiently confident about the object it is observing, along with the object's pose. The introduction of the model-free policy significantly increases the number of episodes where Monty achieves this high-confidence terminal state. In addition, we observe that the total accuracy is improved. This is likely due to Monty encountering features that disambiguate between objects, but which are not visible to the random-walk distant agent.

\textbf{Model-based policies} are believed to be vital to intelligence, yet how to efficiently learn and leverage the necessary representations remains an open problem \parencite{Tolman1948CognitiveMen, schneider2024surprising}. In this work, Monty combines learned internal models with an innate policy for distinguishing ambiguous objects, demonstrating the utility of a model-based policy during sensorimotor inference. In particular, Monty is able to use its internal hypotheses for the most likely objects to identify a highly discriminative action; we call this the hypothesis-testing policy. Intuitively, an LM performs a mental rotation of the two most likely hypotheses, comparing these to find regions where they differ significantly (see Section \ref{sec:model_based_method} for details). Figure \ref{fig:rapid_inference_policy}B demonstrates that the hypothesis-testing policy can provide an additional boost to the model-free policy, improving accuracy and speed of convergence on challenging objects (accuracy 96.4\% with vs. 95.6\% without; median steps to convergence 28 with vs. 30 without). 

Given the already strong performance of the curvature-following, model-free surface agent, the absolute change in accuracy from the addition of the model-based policy is small. However, it is worth examining the principled actions that it enables. Figure \ref{fig:rapid_inference_policy}C-D shows an example of the hypothesis-testing policy used to distinguish two similar objects (the fork and spoon), while Figure \ref{fig:rapid_inference_policy}E-F demonstrates its utility for distinguishing alternative pose hypotheses. Rather than Monty moving blindly, the hypothesis-testing policy enables principled actions that efficiently eliminate ambiguity. It is likely that as the adversarial nature of a task increases, such deliberate movements would become increasingly important. We also highlight that this policy does not need to be relearned for each new object, or be guided by reward signals. Just as Tolman's rats learned the structure of mazes without external rewards \parencite{Tolman1948CognitiveMen}, Monty builds a model of the world (here 3D objects) through open-ended exploration. When it needs to use those models to achieve a task, such as disambiguating object ID and pose, they can be leveraged immediately. 

\subsubsection{Voting for Rapid Inference}

We have so far emphasized the importance of movement when Monty performs inference, given its centrality in natural vision  \parencite{held1963movement, gibson1966senses, Yarbus1967EyeObjects, gilchrist1997saccades}. Computational models of biological vision \parencite{Fukushima1980Neocognitron:Position, Wallis1997InvariantSystem, Riesenhuber1999HierarchicalCortex, Serre2007RobustMechanisms} and their machine learning counterparts \parencite{Lecun1998Gradient-basedRecognition, Krizhevsky2012ImageNetNetworks, Dosovitskiy2021AnScale} typically emphasize the rapid, parallel processing of an entire visual field, without any movement taking place. While we have highlighted that biological systems are constrained to a partial view of the world and must use movement to integrate information, it is also true that the neocortex is able to combine information from multiple sensory inputs to enable rapid recognition \parencite{Thorpe1996SpeedSystem.}. In the most extreme setting, no movement takes place, what is sometimes called `flash' inference \parencite{clay2024thousand}. Importantly, such recognition should use structured representations to enable robustness, just as movement-based recognition does. Monty achieves this through a process termed \textit{voting} \parencite{clay2024thousand}.

\begin{figure}[htbp]
    \centering
    \includegraphics[width=0.8\textwidth]{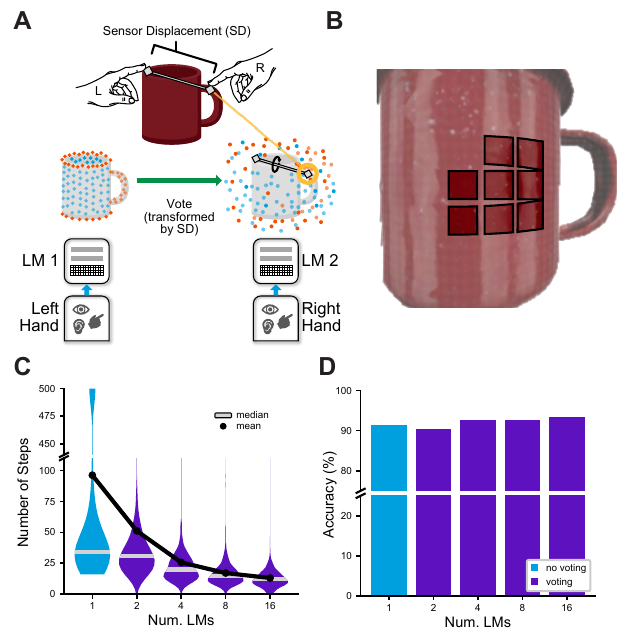}
    \caption{\textbf{Rapid Inference with Voting.} A) LMs receiving information from different sensory patches (such as the index fingers of a left and right hand) can vote with one another to rapidly reach consensus about what object is being sensed. In this diagram, LM1 receives information from the left finger, and hypothesizes that it may be somewhere on the mug's rim. Based on the relative displacement of the fingers, LM1 predicts where LM2 should be on the object, including one hypothesis that corresponds to the handle. These votes are sent as CMP signals that bias the evidence values in LM2. Note that many votes from LM1 to LM2 end up in empty space on LM2's cup model, and these hypotheses are therefore inconsistent with what LM2 is sensing. Votes are sent bidirectionally, but are shown here in one direction for simplicity. B) In these experiments, the different SMs correspond to patches forming a 2D grid in visual space that move together as a unit. Here we show the example of 8 SMs connected to 8 LMs. C) In a multi-LM instance of Monty, $k$ LMs must terminate for the system to converge on a hypothesis. Here we set $k=2$ and vary the total number of LMs in the system. This results in Monty converging in significantly fewer steps as the number of SMs (and associated LMs) scale, while achieving similar classification accuracy (D).}
    \label{fig:rapid_inference_voting}
\end{figure}

For a detailed description of voting, we refer the reader to Section \ref{sec:voting_methods} of our Methods, however Figure \ref{fig:rapid_inference_voting}A demonstrates the intuition behind the algorithm. By sharing information about their hypotheses, LMs can rapidly achieve consensus, minimizing the number of movements required. Importantly, voting does not simply look for agreement between LMs at the coarse level of object ID (e.g., "we are both seeing features found on mugs"), but accounts for the relative displacement between their models and their sensory inputs. This ensures that voting does not function as a bag-of-features operator insensitive to global structure, a limitation frequently associated with deep learning methods \parencite{Brendel2019ApproximatingImageNet, Dosovitskiy2021AnScale, Geirhos2021PartialVision, gavrikovcan}.

As a result of this sensitivity to spatial displacements, voting can take place between two independently moving sensors, such as two fingers, one on each hand \parencite{clay2024thousand}. In this work, we consider the biological analogy of retinal patches that move together, forming a grid of sensory patches in the distant agent (Figure \ref{fig:rapid_inference_voting}B). While the sensors in these grids have fixed relative displacements, they communicate sensed locations in 3D space, the relative displacement ($\prescript{B}{}d_t = \prescript{B}{}x_{t}^{l+1} - \prescript{B}{}{x_t}^{\hat{l}}$) of which is \textit{not} fixed as the sensors move. For example, as Monty moves across the surface of a curved object, a location sensed at a point of steep curvature will be farther away in the depth direction than other points on the grid. As discussed in Section \ref{sec:voting_methods}, Monty's LMs transform votes by relative displacements to ensure these dynamic changes are accounted for.

Figure \ref{fig:rapid_inference_voting}C and D demonstrate the benefits that voting introduces. By enabling multiple LMs to communicate their hypotheses, Monty can converge to a likely representation with far fewer steps, a result that scales with the number of SMs used in the Monty system. Importantly, this increased efficiency does not come at a cost in robustness, with accuracy remaining approximately level.

These results demonstrate that Monty is able to leverage voting, but is not dependent on it. This is analogous to how you can recognize an object by moving a single finger over its surface. The same is true when viewing an image through a narrow aperture, such as a straw. Using all your fingers or having access to a full visual field enables you to recognize these objects more quickly, but it is not required.

We conclude by noting that, although voting supports simultaneous processing of multiple sensory inputs, in general only a fraction of the relevant information in the world can be perceived at a given point in time (see, for example, the scale of the grid in Figure \ref{fig:rapid_inference_voting}B relative to the mug). As such, voting and movement operate side-by-side, complementing rather than replacing one another. This is a deliberate design choice, ensuring that Monty does not adopt the false assumption that all useful information can be simultaneously perceived.

\subsection{Rapid, Continual, and Efficient Learning}
\label{sec:rapid_learning}

Next, we explore Monty's ability to learn given limited training data, as well as its natural support for continual learning. Following this, we will demonstrate that learning in Monty is also computationally efficient.

\subsubsection{Rapid Learning}

\begin{figure}[htbp]
    \centering
    \includegraphics[width=1.0\textwidth]{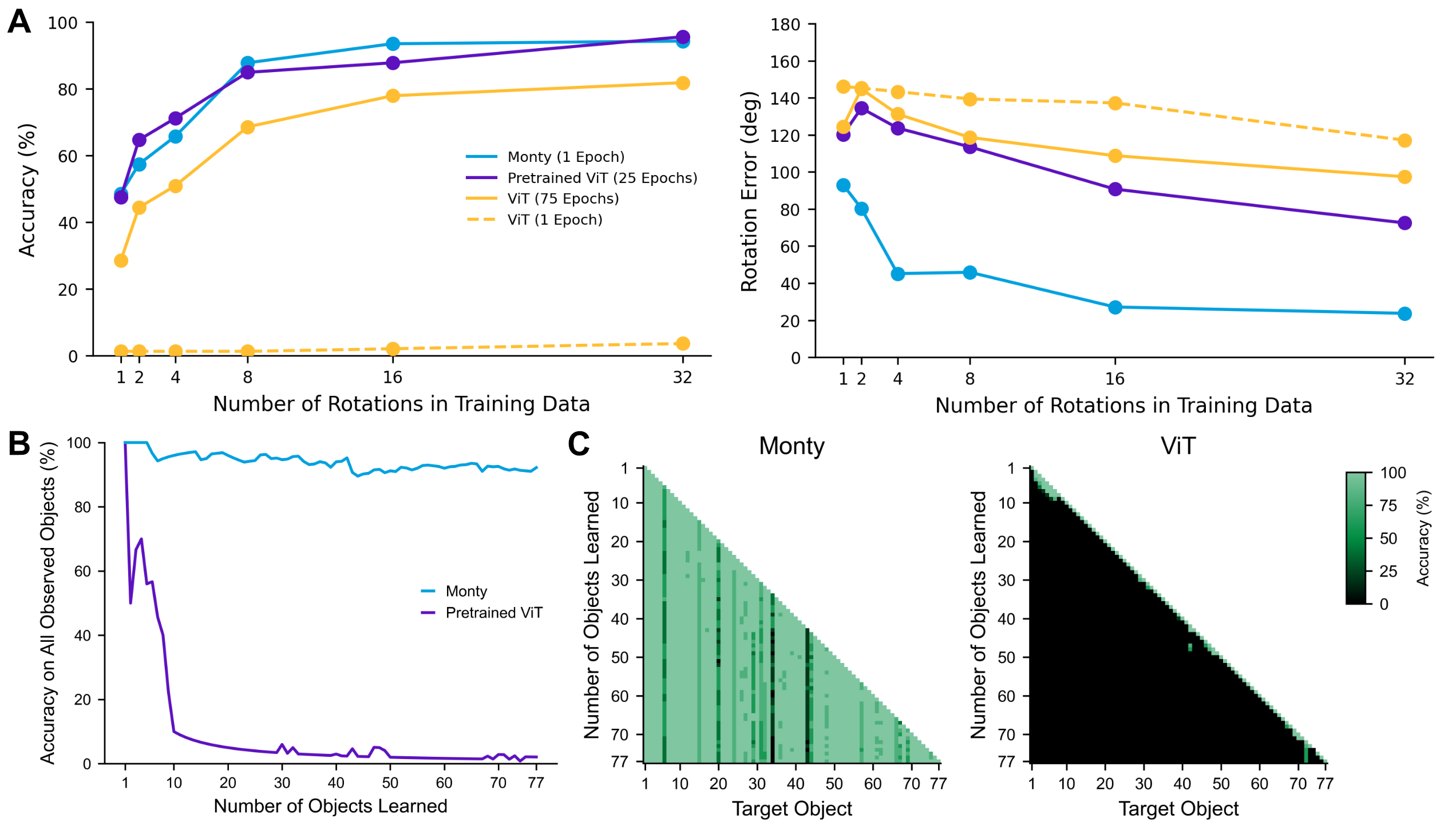}
    \caption{\textbf{Few-Shot and Continual Learning.} A) We evaluate performance after training on all 77 objects, but where each object is presented from a small number of views. We compare to a fine-tuned ViT (previously trained on 14 million images), and a ViT trained from scratch. The ViT trained from scratch receives either 75 epochs (optimal) or 1 epoch (matched to Monty) of exposure to the dataset. Monty learns rapidly through its use of associative connections within reference frames, achieving strong performance in both classification accuracy (left) and pose prediction (right) after limited training. B) We evaluate continual learning performance on a split version of YCB, consisting of 77 "tasks", where 1 object is learned in each. Models are assessed on their ability to classify all objects that have been encountered on the current and previous tasks ("Accuracy on All Observed Objects"). Learning in Monty only modifies local connectivity in the form of associative binding between observed features and the active location in a reference frame. These changes are therefore sparse with respect to the entirety of Monty's representations, conferring robustness to continual learning. C) A breakdown of continual learning performance, showing accuracy on the most recently learned task (diagonal), alongside each of the previously learned tasks (lower-left triangle).}
    \label{fig:rapid_learning} 
\end{figure}

To assess learning under the setting of limited training data, we present Monty with all 77 YCB objects, but vary how many rotations it observes of each object during learning. As these experiments leverage the distant agent, the sensor cannot move around the object. Intuitively, this means that Monty learns the various faces of each object, but only as more rotations are introduced. After 6 rotations, it will have seen all sides of an object corresponding to the faces of a cube, and beyond 14 rotations, it will have seen the additional 8 rotations corresponding to the corners of a cube. During evaluation, objects are presented in 5 novel rotations that differ significantly from the 14 canonical views. Figure \ref{fig:rapid_learning}A demonstrates that after only a handful of such exposures to each object, Monty begins achieving strong classification accuracy and pose estimation (88\% accuracy and 46 degrees mean rotation error after 8 observed views). To put this amount of data into context, 77 objects at 8 rotations is around 600 training samples, 100 times fewer samples than are found in the MNIST dataset \parencite{Lecun1998Gradient-basedRecognition}.

We also compare Monty's performance to vision transformer (ViT) networks \parencite{Dosovitskiy2021AnScale} (further details in Methods Section \ref{sec:vit_methods}). We emphasize that the aim of this work is to demonstrate the capabilities of Monty, the first implementation of a thousand-brains system. The primary purpose is not to argue that Monty, in its current form, is superior to all learning systems. However, it is natural to wonder how deep learning systems would compare in the task setting we consider. For all of the following experiments (unless noted otherwise), we therefore use Monty coupled with a distant agent that can only reorient its sensor from a single, fixed location in space, with no use of the hypothesis-testing policy. This ensures that Monty and the ViTs are similarly unable to reduce self-occlusion of objects. We emphasize, however, that this is an unnatural and limiting condition for Monty, which as a sensorimotor system, excels when it is not subject to such constraints. We also refrain from adding feature noise in any of these experiments, as this would not be directly analogous across the two architecture types.

Figure \ref{fig:rapid_learning}A demonstrates that a ViT network trained from scratch lags significantly behind Monty after exposure to such a small amount of training data. Only a ViT network that has been pre-trained on 14 million RGB images achieves similar learning efficiency to Monty, and importantly, only on the in-domain task of object classification. For the out-of-domain task of pose prediction (Figure \ref{fig:rapid_learning}A right), the pre-trained ViTs features are insufficient to enable rapid generalization, and we were often unable to identify hyperparameters (Methods Section \ref{sec:vit_methods}) that improved rotation error when the network received fewer than 32 rotations in the training data.

We also consider a network that is trained from scratch with only a single epoch of training (i.e., only sees each object in each rotation once). This is the only ViT network that receives the same amount of data as Monty. While Monty, like humans \parencite{lake2011one} can learn from single exposures, these results demonstrate that the ViT never achieves accuracy significantly above chance.

What underlies Monty's capabilities in rapid learning? We argue that the key element is the use of local, associative learning together with structured reference frames. When combined, a single exposure to an object is enough to rapidly lay down representations that can be leveraged for inference. The natural symmetry of objects further aids generalization. For example, after Monty has learned one face of an object, this representation will generalize to any other sides of the object that share its appearance. This is possible even when the face is oriented in an unfamiliar manner. Recall that Monty achieves rotation invariance by directly inferring the possible rotation, transforming incoming sensations and displacements based on this inferred pose. As such, we observe that Monty achieves approximately 50\% classification accuracy (chance would be 1.3\%) after observing only a single view of each object. This contrasts with an accuracy of approximately 30\% in the ViT trained from scratch. These results provide a compelling example of out-of-distribution (OOD) generalization in Monty, one that cannot be explained by merely interpolating learned examples.

\subsubsection{Continual Learning}

Next, we evaluate Monty's capabilities in the domain of continual learning (CL). Humans are able to learn new skills and knowledge throughout their lifetime without displaying the phenomenon of catastrophic forgetting \parencite{McCloskey1989CatastrophicProblem}. To evaluate Monty's CL capabilities, we split the YCB dataset into a set of 77 `tasks' \parencite{Zenke2017ContinualIntelligence}, with one object in each. We train on each of these tasks in sequence; while the system is learning to recognize one object, it does not observe any instances of other objects. We then evaluate accuracy by assessing performance on all objects that have been encountered so far. To score well, Monty must learn the new object in the $n$th task, while retaining knowledge about objects in the previous $n-1$ tasks.

Figure \ref{fig:rapid_learning}B and C demonstrate that Monty maintains strong performance throughout all 77 tasks, displaying only a minor degradation due to interference between similar objects as a greater proportion of the YCB dataset is learned. By updating the representation that is currently active ($(\prescript{M}{}{x_i}, \prescript{M}{S}{\mathbf{R}_i}, n_i) \in \mathcal{M}^m$), and only this representation, Monty's associative binding can be viewed as a form of local, sparse weight formation. As such, other learned representations remain unaffected, supporting CL.

We once again compare to ViT networks under the same training paradigm, selecting the higher-performing pre-trained ViT. As observed in Figure \ref{fig:rapid_learning}, the ViT displays catastrophic forgetting \parencite{McCloskey1989CatastrophicProblem}, overwriting weight changes that supported earlier tasks. Unlike Monty's use of local learning in a reference frame, learning in the ViT relies on global changes to all of the weights of the network, modifying these in ways that impact its ability to use the same weights for earlier tasks.

We note that our task setup extrapolates the typical approach taken for CL datasets, namely splitting the dataset into a set of tasks, with a set of objects in each \parencite{Zenke2017ContinualIntelligence}. We examine our one-object-per-task condition because it is a learning paradigm that must be supported to accommodate the statistics of the natural world. For example, an animal learning about edible fruits will typically not encounter these as a batched input that can be held side-by-side. Instead, objects in the world are spatially and temporally correlated \parencite{Condit2000SpatialSpecies}. It is notable that humans benefit from learning when like inputs are clustered in time, rather than interleaved as required by deep learning systems \parencite{Flesch2018ComparingMachines}.

We also highlight that when training the ViT, we do not mask the cross-entropy loss for previous tasks, as often employed in CL \parencite{boschini2022class}. When multiple objects are learned in a batch-like task, such masking ensures the network is not punished for predicting previously seen (but currently irrelevant) objects. In the task setting we consider however, such masking would result in only the logit for the current (singular) object having an output. Applying softmax to a single unit and then back-propagating would not provide a meaningful gradient for the network to learn, underscoring the reliance of deep learning methods on contrastive signals to develop representations.

A possible objection to our findings is that Monty, in its current form, continuously expands its representational capacity when learning. In particular, when a new object is encountered, Monty initializes a new reference frame for the object, and when learning a new point on an object, a new value is stored in memory. However, prior work has demonstrated that a reference-frame-based model with Hebbian learning is robust to catastrophic forgetting, even when the system has a fixed representational capacity \parencite{Leadholm2021GridRecognition}. Furthermore, the ViT model we compare to has approximately 86 million parameters \parencite{Dosovitskiy2021AnScale}. In contrast, Monty has approximately 4 million parameters after learning on all of the YCB dataset. As such, it is the local nature of Monty's learning, rather than the expansion of its representational capacity, that is key to its CL capabilities.

\subsubsection{Computational Efficiency}

As a final measure of learning in Monty, we examine its computational efficiency, quantified via floating point operations (FLOPs). We train Monty in our standard setup (77 YCB objects, 14 rotations each), and track how many FLOPs are performed throughout the entire learning process. To put the result into context, we once again compare to ViT networks, this time considering both the from-scratch network, as well as the pre-trained network. For the latter, we include FLOPs for fine-tuning, as well as estimated FLOPs for the pre-training stage (details in Methods Section \ref{sec:flops_methods}).

\begin{figure}[htbp]
    \centering
    \includegraphics[width=1.0\textwidth]{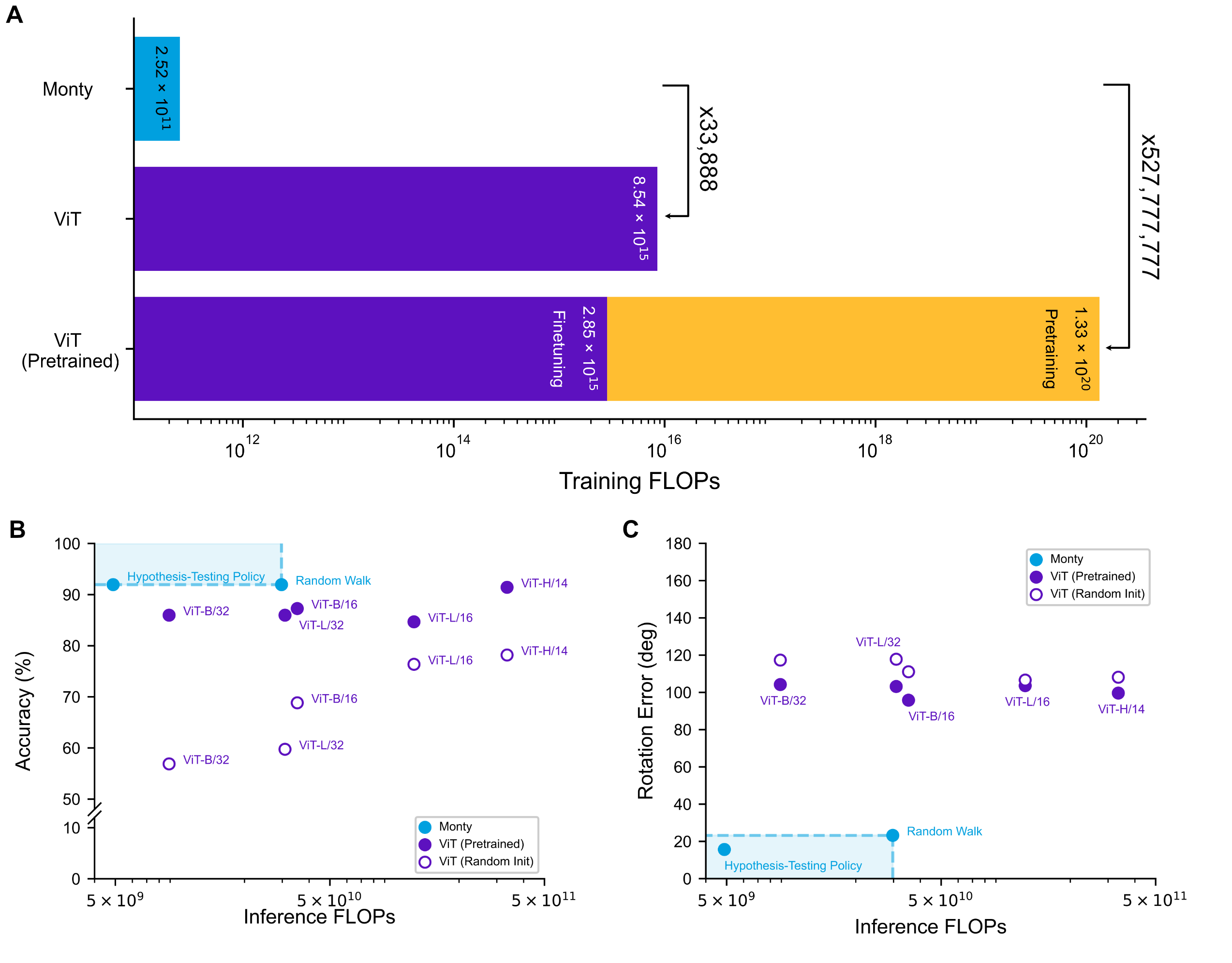}
    \caption{\textbf{Computationally Efficient Learning and Inference.} A) During learning, Monty only modifies representations in the current reference frame, at the current location where information is sensed. This results in a massive reduction in the Floating Point Operations (FLOPs) required for learning when compared to the global gradient calculation and weight updates required in deep learning architectures. This difference is already apparent when comparing to the from-scratch ViT, despite the latter underperforming on accuracy metrics. In the case of the ViT that has undergone additional pre-training, the difference is a factor of nine orders of magnitude. B) We visualize the average inference FLOPs associated with processing a single object. Even at inference, Monty compares favorably to deep learning architectures. In our ViT comparisons, we restrict Monty to viewing only one side of the object so that the data the models receive is as comparable as possible. However, if Monty makes full use of its sensorimotor capabilities and efficiently explores the object, there is a significant drop in the number of FLOPs required to achieve its performance.}
    \label{fig:efficient_learning}
\end{figure}

Figure \ref{fig:efficient_learning}A demonstrates that Monty uses several orders of magnitude fewer FLOPs than the ViT networks. Compared to the from-scratch network, Monty requires approximately 34,0000$\times$ fewer FLOPs, despite Monty demonstrating stronger accuracy in the few-shot learning setting (Figure \ref{fig:rapid_learning}A). Compared to the pre-trained ViT, the only network we examine that matches Monty's object classification accuracy, Monty requires approximately 528,000,000$\times$ fewer FLOPs for training. Such computationally efficient learning is critical given the importance of life-long learning in intelligent systems, complementing the rapid and continual learning demonstrated in the previous results (Figure \ref{fig:rapid_learning}B).

As an additional dimension, we consider FLOPs used during inference. Given the large search space of hypotheses that Monty considers during inference, it is natural to wonder whether it significantly underperforms ViT networks in this setting. However, Figure \ref{fig:efficient_learning}B demonstrates that Monty compares favorably to the ViT networks, achieving higher accuracy as a function of FLOPs. We note that Monty currently suffers from the issue that inference FLOPs scales linearly with the number of known models. However, this would be mitigated by future learning rules that merge existing models over time, including hierarchically decomposing objects to enable greater model re-use.

Finally, performing inference from a single fixed vantage point is unnatural for a sensorimotor system such as Monty. We therefore include a condition where Monty is able to move around objects using the hypothesis-testing policy. In this setting, Monty gains a significant edge over the ViT networks when trading off accuracy and efficiency (Figure \ref{fig:efficient_learning}B), underscoring the more general potential of sensorimotor systems.

\section{Conclusion}

Building off of prior neuroscience theory, \textcite{clay2024thousand} proposed the concept of a thousand-brains system. This architecture was presented as a new form of sensorimotor AI, one that might carry a variety of desirable properties for intelligent systems. However, these performance characteristics had not been quantified. We set out to evaluate the properties of Monty, an open-source implementation of a thousand-brains system. Our experiments, first and foremost, support the claim of robust inference owing to Monty's structured representations. Leveraging intelligent motor policies, along with a multi-LM voting algorithm, Monty also demonstrates rapid inference. Finally, we observed that the use of local, associative learning within reference frames enables rapid, continual, and computationally efficient learning. We emphasize that this constellation of properties did not emerge from focusing on one of the many open problems within machine learning, such as continual learning or shape bias. Rather, they emerged naturally through the development of a sensorimotor system informed by neuroscience theory, which was in turn informed by evidence from neurobiology \parencite{Hawkins2019ANeocortex, hawkins2021thousand, Hawkins2025}.

This work forms part of an ambitious and long-term effort to develop fundamentally intelligent systems, an effort known as the Thousand Brains Project \parencite{clay2024thousand}. We recognize that this research is still at an early stage, and that the current version of Monty represents an imperfect instance of its final vision. For example, we have limited our evaluations to the context of 3D object perception, the first use case that Monty's implementation supports. We have also not considered important aspects necessary for an intelligent system, such as modeling objects that can move and display complex behaviors, representing compositional objects through a hierarchy of LMs \parencite{Hawkins2025}, or how to coordinate an action policy that changes the state of the external world. Finally, Monty is designed with unsupervised learning at its core, but exploring this paradigm was beyond the scope of the present work.

As the capabilities of thousand-brains systems grow, we anticipate a variety of benefits in downstream applications. The world is filled with tasks that require sensorimotor intelligence capable of learning robustly and quickly from limited and unlabeled data. Such tasks can be found in settings as diverse as controlling agricultural pests, maintaining infrastructure in the renewable energy sector, and providing medical ultrasound in resource-limited settings.

These are exciting avenues for future research. Until such work can be carried out, we believe the present study serves as a useful demonstration of the underlying potential of thousand-brains systems, as well as sensorimotor learning more generally.

\section{Acknowledgments}

We would like to thank the following individuals for invaluable discussions on the Thousand Brains theory and Monty concepts: Subutai Ahmad, Heiko Hoffmann, Kevin Hunter, and Will Warren. In addition to such contributions to discussions, we would like to thank the following individuals for their contributions to the Monty code base: Ben Cohen, Jad Hanna, Abhiram Iyer, Ramy Mounir, Luiz Scheinkman, Philip Shamash, Tristan Slominski, and Lucas Souza.

\printbibliography

\newpage

\section{Appendix}

\subsection{Summary of Mathematical Notation}

Below, we include Tables \ref{tab:math_notation} and \ref{tab:math_notation_2} to summarize the mathematical notation we use.

\begin{table}[h!]
\centering
\renewcommand{\arraystretch}{1.2}
\begin{tabular}{ll}
\toprule
\textbf{Symbol} & \textbf{Description} \\
\midrule
\multicolumn{2}{l}{\textit{Models and Learning}} \\
$B$ & Shared, body-centric coordinate system \\
$m$ & Object identity label \\
$M$ & Coordinate system of an object model; implicitly for object $m$ unless specified \\
$S$ & Coordinate system of a local sensory observation (e.g., surface-patch) \\
$\prescript{B}{S}{\mathbf{R}}_{t}$ & Rotation of $S$ w.r.t.\ $B$ at step $t$ \\
$\prescript{B}{}{x_t}$ & Location of a sensory observation (e.g., surface-patch) in $B$ at step $t$ \\
$n_t$ & Non-pose features (HSV, curvature magnitude, etc.) at step $t$ \\
$\phi$ & Cortical Messaging Protocol (CMP) message \\
$\prescript{B}{M}{\mathbf{R}}$ & Rotation of object frame $M$ w.r.t.\ $B$ \\
$\mathcal{M}^m$ & Set of learned  representations for object $m$, also referred to as the `model' for object $m$ \\
$\prescript{M}{}{x_t}$ & Active location estimate at step $t$ in internal, object reference frame $M$\\
$\prescript{M}{}{x_i}$ & Location of learned representation $i$ in $M$ \\
$\prescript{M}{S}{\mathbf{R}}_{i}$ & Local rotation of learned representation $i$ in $M$ \\
$n_i$ & Non-pose features of learned representation $i$\\
$v_t$ & Movement vector between steps $t$ and $t\!-\!1$ \\
\midrule
\multicolumn{2}{l}{\textit{Inference}} \\
$\mathcal{H}_t^{l}$ & Set of $K$ hypotheses held by LM $l$ at step $t$ \\
$\prescript{M}{}{x_{k,t}}$ & Hypothesized location in $M$ for hypothesis $k$ at step $t$ \\
$\prescript{B}{M}{\mathbf{R}_k}$ & Hypothesized rotation of $M$ for hypothesis $k$ at step $t$ \\
$e_{k,t}$ & Evidence score for hypothesis $k$ at step $t$ \\
$\mathcal{X}^m$ & Set of all learned locations for object $m$ \\
$\mathcal{N}_\varepsilon$ & Set of of points within $\varepsilon$ neighborhood\\
$D(\cdot,\cdot)$ & Distance function used to compare features \\
$\Delta e^{\mathbf{R}}_{k,t}$ & Evidence change from pose-feature comparison \\
$\Delta e^{n}_{k,t}$ & Non-negative evidence change from non-pose features comparison \\
$\theta_\text{converge}$ & Evidence-gap threshold for LM convergence \\
$\theta_\text{update}$ & Evidence-gap threshold for whether to update a hypothesis \\
$k^*$ & Index of the most-likely hypothesis (MLH) \\
$\mathcal{R}^{m}$ & Set of high-evidence rotation-hypotheses for object $m$ \\
$\tau_{\text{sym}}$ & Consecutive-step counter for symmetry detection \\
$\theta_{\text{sym}}$ & Threshold on $\tau_{\text{sym}}$ for declaring symmetry \\
$D_{\text{geo}}$ & Geodesic distance between two rotations \\
$E^{\text{rot}}$ & Rotation error relative to ground truth \\
\bottomrule
\end{tabular}
\vspace{10pt}
\caption{\textbf{Table of Core Mathematical Notations}}
\label{tab:math_notation}
\end{table}

\begin{table}[h!]
\centering
\renewcommand{\arraystretch}{1.2}
\begin{tabular}{ll}
\toprule
\textbf{Symbol} & \textbf{Description} \\
\midrule
\multicolumn{2}{l}{\textit{Movement and Policies}} \\
$a_t$ & Motor action executed at step $t$ \\
$q$ & Object identity label of the second-most likely object \\
$h^*$ & MLH index associated with object $q$ \\
$\prescript{B}{}{\mathcal{X}^m}$ & Learned locations for $m$ offset by MLH and transformed into $B$ \\
$i^*$ & Index of model point maximizing hypothesis discrimination \\
\midrule
\multicolumn{2}{l}{\textit{Voting}} \\
$\hat{l}$ & Index of learning module sending votes \\
$d_t$ & Instantaneous displacement between two SM observations \\
$\prescript{M}{}{\hat{x}_{k,t}}$ & Location of hypothesis $k$, sent as vote by LM $\hat{l}$ \\
\midrule
\multicolumn{2}{l}{\textit{Deep Neural Networks}} \\
$L_{\text{cls}}$ & Cross-entropy loss for classification \\
$L_{\text{rot}}$ & Geodesic loss for rotation predictions \\
$\lambda$ & Weighting factor between classification and rotation loss \\
\bottomrule
\end{tabular}
\vspace{10pt}
\caption{\textbf{Table of Additional Mathematical Notations}}
\label{tab:math_notation_2}
\end{table}

\newpage
\subsection{Noise Parameters for Robustness Experiments}

Below, we summarize the noise parameters used in our robustness evaluations.

\begin{table}[h]
\centering
\begin{tabular}{lllll}
\toprule
\textbf{Eval. Condition:} & \textbf{Noise Only} & \textbf{New Rotations} & \textbf{Combined} & \textbf{New Color} \\
\midrule
\multicolumn{4}{l}{\textit{Noise Parameters}} \\
Location Noise (G) & 0.002 m & - & 0.002 m & 0.002 m \\
Hue Noise (G) & 0.1 & - & 0.1 & HSV := (0.667, 1.0, 1.0) \\
Pose Vector Noise (G) & 2.0° & - & 2.0° & 2.0° \\
Curvature Log Noise (G) & 0.1 & - & 0.1 & 0.1 \\
Non-Unique Pose (BSC) & 0.01 & - & 0.01 & 0.01 \\
\midrule
\multicolumn{4}{l}{\textit{Rotation Parameters}} \\
Test Views & = Train Views & $\prescript{B}{M}{\mathbf{R}}^{\text{gt}} \sim \mathbb{U}(\mathrm{SO}(3))$ & $\prescript{B}{M}{\mathbf{R}}^{\text{gt}} \sim \mathbb{U}(\mathrm{SO}(3))$ & $\prescript{B}{M}{\mathbf{R}}^{\text{gt}} \sim \mathbb{U}(\mathrm{SO}(3))$\\
\bottomrule
\end{tabular}
\vspace{10pt}
\caption{\textbf{Configuration Details for Robustness Experiments}: Noise parameters are either the standard deviation of a Gaussian distribution (marked G), or the probability of flipping a boolean variable according to a binary symmetric channel model ($\text{BSC}_p$), marked BSC. \textit{Hue Noise}: Perturbation of hue values, with the full hue spectrum scaled to [0, 1.0]. \textit{Pose Vector Noise}: Degrees by which the vectors defining a given $\prescript{B}{S}{\mathbf{R}}_t$ are perturbed. \textit{Curvature Log Noise}: Perturbation of measured curvature magnitudes, in log-space. \textit{Non-Unique Pose}: A pose observation $\prescript{B}{S}{\mathbf{R}}_t$ may not have a unique definition if the principal curvature directions are undefined (e.g., on a flat surface). Monty's SMs estimate whether this is true or not for downstream processing; this noise randomly flips the boolean result with probability $p$. \textit{New Color}: All HSV values, irrespective of the observation or object, are set to hue: 0.667, saturation: 1.0, and value: 1.0.}
\label{tab:robustness_config}
\end{table}

\end{document}